\pdfoutput=1
\documentclass{article}

\PassOptionsToPackage{numbers,compress}{natbib}
\PassOptionsToPackage{table}{xcolor}
\usepackage[preprint]{neurips_2026}
\usepackage[utf8]{inputenc}
\usepackage[T1]{fontenc}
\usepackage{hyperref}
\usepackage{url}
\usepackage{booktabs}
\usepackage{enumitem}
\usepackage{amsmath,amssymb,amsfonts}
\usepackage{nicefrac}
\usepackage{algorithm }
\usepackage{booktabs}
\usepackage{tabularx}
\usepackage{placeins}
\usepackage{algpseudocode}
\usepackage{xcolor}
\usepackage{colortbl} 
\usepackage{booktabs}
\usepackage{multirow}
\usepackage{threeparttable}
\usepackage{tikz}
\usepackage{microtype}
\usepackage{booktabs}
\usepackage{array}
\usepackage{graphicx}
\usepackage{multirow}
\usepackage{makecell}
\usepackage{xcolor}
\usepackage{threeparttable}
\usepackage{subcaption}
\newcommand*\circled[1]{\tikz[baseline=(char.base)]{\node[shape=circle,fill=black,text=white,draw,inner sep=.1pt] (char) {#1};}}

\makeatletter
\renewcommand{\@notice}{}
\makeatother
\title{EGL-SCA: Structural Credit Assignment for Co-Evolving Instructions and Tools in Graph Reasoning Agents}

\author{
\textbf{Zike Yuan}$^{1,2}$\thanks{Equal contribution.},
\textbf{Yukun Cao}$^{3}$\footnotemark[1],
\textbf{Han Zhang}$^{1,2}$,
\textbf{Jianzhi Yan}$^{1,2}$,
\textbf{Le Liu}$^{1,2}$,
\textbf{Cai Ke}$^{1,2}$,\\
\textbf{Yue Yu}$^{2}$,
\textbf{Hui Wang}$^{2}$,
\textbf{Ming Liu}$^{1,2}$,
\textbf{Bing Qin}$^{1,2}$\\[0.3em]
{\normalfont $^{1}$Harbin Institute of Technology, Shenzhen, China}\\
{\normalfont $^{2}$Peng Cheng Laboratory, Shenzhen, China}\\
{\normalfont $^{3}$School of Computer Science and Technology, Xidian University}\\[0.2em]
{\normalfont \texttt{\{yuanzk,yanjzh,liul07,wangh06\}@pcl.ac.cn}}\\
{\normalfont \texttt{HanlardResearch@gmail.com}; \texttt{kecai@stu.hit.edu.cn}}\\
{\normalfont \texttt{caoyukun@xidian.edu.cn}; \texttt{\{mliu,qinb\}@ir.hit.edu.cn}}
}

\begin{document}

\maketitle
\nocite{alphaevolve2025,aflow2024,brown2020language,cao2024graphinsight,chen2024graphwiz,fernando2023promptbreeder,gepa2025,zhang2023graphmeetsllms,graphtoolformer2023,luo2024graphinstruct,opsahlong2024mipro,paranjape2023art,romera2023funsearch,schick2023toolformer,shang2024agentsquare,shinn2023reflexion,tang2024grapharena,wang2023nlgraph,wei2022chain,xu2025graphomni,yao2023react,yuan2024gracore,yuan2025magts,zhao2023expel,gao2023pal,chen2023programofthoughts,deb2002nsga2}

\begin{abstract}
Graph reasoning agents operating from natural-language inputs must solve a coupled problem: they must reconstruct a structured graph instance from text, decide whether existing computational assets are sufficient, interact with tools under a strict execution protocol, and satisfy an external verifier that checks structured correctness rather than textual plausibility. Existing approaches usually improve either the instruction side or the tool side in isolation, which leaves unclear what should be updated after failure. We propose EGL-SCA, a verifier-centric dual-space framework that models a graph reasoning agent using two collaborative components: an instruction-side policy space for reasoning strategies, and a tool-side program space for executable algorithmic tools. Our central mechanism is {\it structural credit assignment}, which maps trajectory evidence to conditional updates, precisely routing failures to either prompt optimization or tool synthesis and repair. To provide sufficient learning signals for dual-space adaptation, we introduce a training distribution stratified by task family, coupled with a Pareto-style retention strategy to balance success, generality, and parsimony. Experiments on four graph reasoning benchmarks show that EGL-SCA achieves a state-of-the-art 92.0\% average success rate. By effectively co-evolving instructions and tools, our framework significantly outperforms both pure-prompting and fixed-toolbox baselines.
\end{abstract}

\section{Introduction}

Graph reasoning remains a persistent weakness of large language model (LLM) agents when tasks are presented as natural-language problem statements rather than as canonical graph objects~\citep{wang2023nlgraph,luo2024graphinstruct,tang2024grapharena,yuan2024gracore}. In realistic settings, an agent must first recover a structured instance from text: it must identify the node set, edge relations, directionality, weights, query variables, and side constraints, and then decide how that reconstructed object should be solved. This difficulty is qualitatively different from unconstrained text generation. A model may produce a fluent answer while solving the wrong graph, selecting an inappropriate algorithmic routine, or violating an execution protocol required by an external verifier~\citep{cao2024graphinsight,wang2023nlgraph}. As a result, the dominant errors in graph reasoning are often not merely logical mistakes in the final answer; they are upstream errors in structured reconstruction, solver choice, or executable interaction.

\begin{figure*}[!t]
  \centering
  \includegraphics[width=0.95\textwidth]{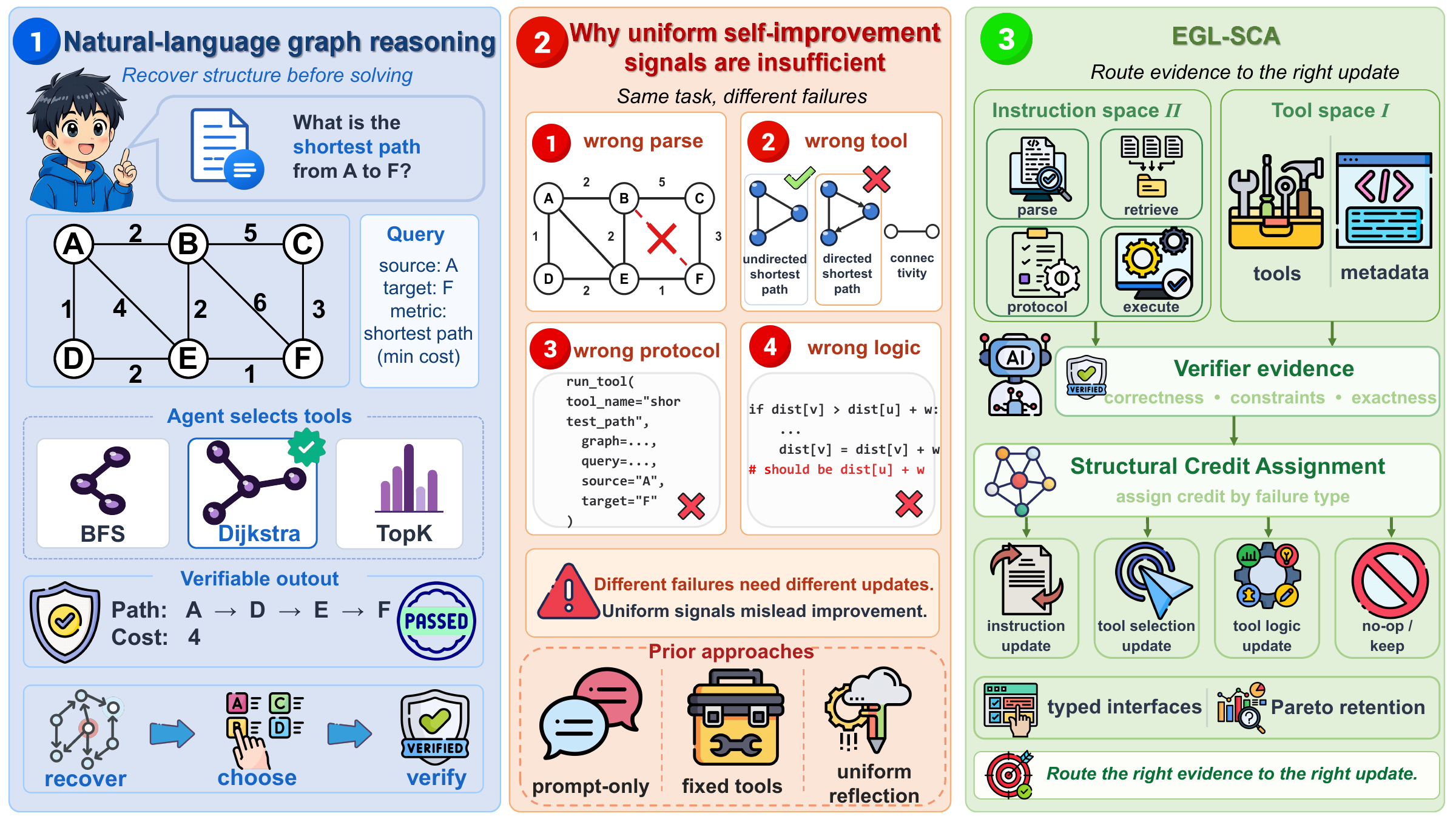}
  \caption{
Motivation: EGL-SCA routes graph-reasoning failures to the component that caused them.
  }
  \label{fig:intro_overview}
\end{figure*}

This observation exposes a limitation shared by several existing lines of work. \circled{1} Prompt-only reasoning methods~\citep{brown2020language,wei2022chain} can produce plausible solutions on small graph instances, but they are brittle when exact structural recovery and verifiable outputs are required. \circled{2} Tool-using agents~\citep{yao2023react,schick2023toolformer,paranjape2023art,gao2023pal,chen2023programofthoughts} improve execution, yet they usually assume a fixed toolbox and therefore cannot accumulate new graph-specific algorithmic tools during training. \circled{3} Reflection-based self-improvement methods~\citep{shinn2023reflexion,zhao2023expel} can refine language behavior through feedback, but they typically treat failure as a single undifferentiated signal. In graph reasoning, such undifferentiated feedback is inadequate. A failure caused by an incorrectly reconstructed constraint should improve the parsing behavior; a failure caused by choosing an unsuitable solver should improve retrieval or selection; and a failure caused by a tool-logic error should improve the tool itself. Treating all of these failures as the same learning event leads to misdirected updates and unstable training.

We argue that self-improving graph reasoning requires \emph{conditional adaptation over structured solve trajectories}. The central question is not simply how to improve an agent after failure, but \emph{what should be improved, and why}. To address this question, we conceptualize a graph reasoning agent as a dual-space system composed of two distinct but complementary halves: \textbf{an instruction-side policy space and a tool-side program space}. The instruction side governs how the agent reconstructs task structure, retrieves relevant capabilities, follows an interaction protocol, and formats executable inputs. The tool side contains reusable algorithmic procedures together with metadata describing their applicability. Under this view, learning no longer amounts to editing a single prompt or growing an unstructured toolbox; it becomes a problem of coordinating updates between two coupled spaces that fail for different reasons.

Our proposed framework, \textbf{EGL-SCA} (\textbf{E}volving \textbf{G}raph \textbf{L}earning via \textbf{S}tructural \textbf{C}redit \textbf{A}ssignment), resolves this coordination problem through its core mechanism of \emph{structural credit assignment}. Given a solve trajectory, task-level diagnostics, and verifier feedback, EGL-SCA infers which component of the agent was primarily responsible for failure. It distinguishes between failures attributable to the instruction side, failures caused by incorrect tool selection, failures caused by tool-logic errors, and cases in which no update is necessary. Moreover, instruction-side failures are further localized to more specific subproblems. EGL-SCA therefore treats failure as a trajectory-level object rather than a terminal binary label.

To retain the highest-quality adaptations, EGL-SCA captures intermediate behavior by retaining policy pairs according to a Pareto-style criterion over success, generality, protocol reliability, and parsimony~\citep{deb2002nsga2}. Finally, because learning dynamics depend heavily on the environment, we expose the necessary learning pressure through a family-aware training distribution that varies graph scale, structural regime, active constraints, and exactness requirements. 

Extensive experiments validate our framework. Across four graph benchmarks, EGL-SCA achieves a state-of-the-art \textbf{92.0\%} average success rate, decisively outperforming pure-prompting (e.g., ExpeL at 20.6\%) and fixed-toolbox agents (e.g., AgentSquare at 87.8\%). Furthermore, our mechanism prevents blind mutations, synthesizing a highly reusable, parsimonious toolbox while maintaining an exceptional protocol reliability of 0.936.

\textbf{Main Contributions:} \circled{1} We frame graph reasoning as a \textbf{dual-space learning system}, explicitly decoupling language-based reasoning strategies from executable operations. \circled{2} We introduce \emph{structural credit assignment} to precisely diagnose reasoning failures, routing feedback to either optimize prompts or repair tools. \circled{3} We design a \textbf{family-aware training curriculum} and a \textbf{Pareto-style retention} scheme to ensure the agent learns robust, general, and parsimonious skills. \circled{4} \textbf{EGL-SCA achieves a SOTA 92.0\% success rate} across four graph benchmarks, proving that co-evolving instructions and tools effectively breaks the bottleneck of current LLM reasoning.

\section{Related Work}

\paragraph{Prompting and policy improvement.}
General prompting methods (e.g., few-shot~\citep{brown2020language}, chain-of-thought~\citep{wei2022chain}) improve reasoning by altering input formats without updating model parameters. To further refine language policies, feedback-driven methods like Reflexion~\citep{shinn2023reflexion} and ExpeL~\citep{zhao2023expel} leverage verbal feedback and accumulated experience, while frameworks such as Promptbreeder~\citep{fernando2023promptbreeder}, MIPRO~\citep{opsahlong2024mipro}, and GEPA~\citep{gepa2025} treat prompts as optimizable objects via search and validation. 

\paragraph{Tool use, workflows, and program search.}
Tool-augmented agents (e.g., ReAct~\citep{yao2023react}, Toolformer~\citep{schick2023toolformer}, ART~\citep{paranjape2023art}) interleave language reasoning with external API calls, whereas workflow-level systems like AFlow~\citep{aflow2024} and AgentSquare~\citep{shang2024agentsquare} automate the design of multi-step agent structures. In executable domains, evaluator-guided search algorithms (e.g., FunSearch~\citep{romera2023funsearch}, AlphaEvolve~\citep{alphaevolve2025}) have also proven effective for discovering strong programmatic procedures through iterative proposal and verifier selection.

\paragraph{Graph reasoning and graph-specialized agents.}
Recent benchmarks (NLGraph~\citep{wang2023nlgraph}, GraphInstruct~\citep{luo2024graphinstruct}, GraphArena~\citep{tang2024grapharena}, GraCoRe~\citep{yuan2024gracore}) expose a persistent gap between fluent text generation and exact graph computation. This difficulty is further exacerbated by representational challenges, such as positional bias in serialized graph descriptions~\citep{zhang2023graphmeetsllms, cao2024graphinsight}. To narrow this gap, specialized methods rely on prompt augmentation (Graph-ToolFormer~\citep{graphtoolformer2023}), instruction tuning (GraphWiz~\citep{chen2024graphwiz}), or multi-agent decomposition (MA-GTS~\citep{yuan2025magts}). However, these approaches typically operate within fixed toolboxes or tool pipelines, or optimize language policies in isolation. In contrast, EGL-SCA jointly evolves instruction-side parsing strategies and executable graph tools under verifier feedback, utilizing structured credit assignment to precisely repair failures in protocol, tool choice, or tool-logic errors.

\section{Method}
\begin{figure*}[!t]
  \centering
  \includegraphics[width=\textwidth]{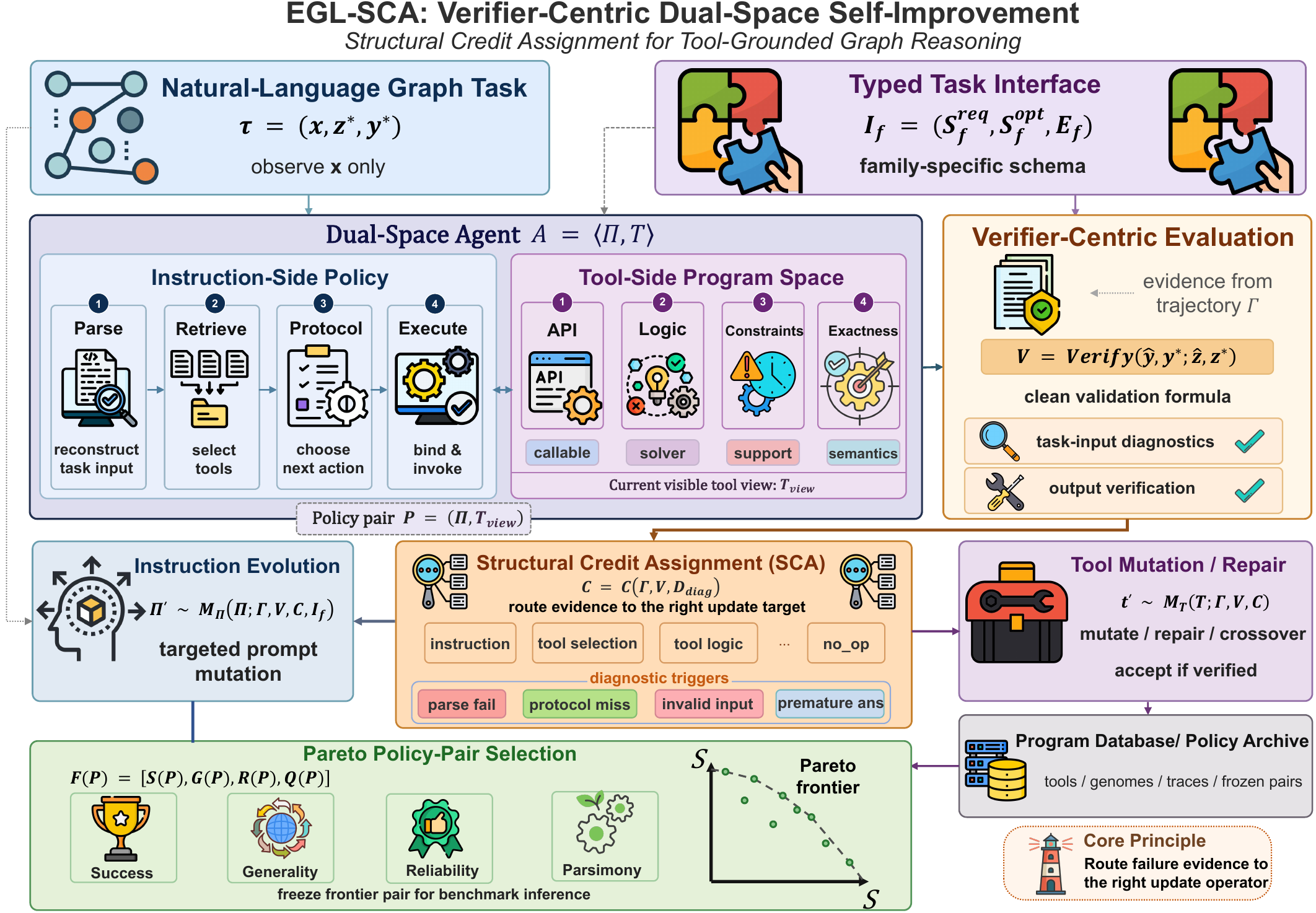}
  \caption{
EGL-SCA overview: verifier evidence coordinates updates between instructions and tools.
  }
  \label{fig:method_overview}
\end{figure*}

\subsection{Problem Setup}
We study graph reasoning tasks expressed in natural language. Each task instance is a tuple $\tau=(x,z^\ast,y^\ast)$, where $x$ is the textual problem statement, $z^\ast$ is the canonical structured task object, and $y^\ast$ is the reference answer. We further write $z^\ast=(g^\ast,q^\ast,c^\ast)$, where $g^\ast$ is the graph structure, $q^\ast$ is the query object, and $c^\ast$ denotes task-specific constraints. At inference time, the agent observes only $x$, not $z^\ast$. The task is therefore not to run an algorithm over a given graph object, but to first reconstruct a candidate structured object $\hat z=(\hat g,\hat q,\hat c)$ from language and then produce an answer $\hat y$ based on that reconstruction. See Appendix~\ref{app:algorithm_sca} for a symbol summary.

A single solve attempt induces a trajectory
\[
\Gamma=(s_1,a_1,o_1,\ldots,s_T,a_T,o_T),
\]
where $s_t$ is the internal state at step $t$, $a_t$ is the chosen action, and $o_t$ is the resulting observation. The trajectory may include task reconstruction, capability retrieval, candidate-tool proposal, tool execution, and verifier-related protocol interactions. The external verifier judges structured correctness rather than textual plausibility, so a failure may arise even when $\hat y$ appears linguistically reasonable. Our goal is therefore not only to maximize end-task success, but also to identify which component first diverged from a valid solution path and should be updated.

\subsection{Dual-Space Agent Representation}
We represent the agent as a dual-space system $A=\langle \Pi,\mathcal{T}\rangle$, where $\Pi$ is an instruction-side policy space and $\mathcal{T}$ is a tool-side program space. The instruction side governs the full decision process from natural language to executable behavior. To make failure attribution explicit, we conceptually decompose it as
\[
\Pi=(\Pi_{\mathrm{parse}},\Pi_{\mathrm{retrieve}},\Pi_{\mathrm{protocol}},\Pi_{\mathrm{execute}}),
\]
where $\Pi_{\mathrm{parse}}$ reconstructs a structured task object from text, $\Pi_{\mathrm{retrieve}}$ determines which tools, documents, or prior assets are relevant, $\Pi_{\mathrm{protocol}}$ selects the next admissible transition in the solve process, and $\Pi_{\mathrm{execute}}$ constructs schema-valid executable inputs and formats outputs correctly. This decomposition does not require four separate models; its purpose is to provide a structural coordinate system for credit assignment and routed updates. In natural-language graph reasoning, many failures arise not because the model ``cannot reason'' in the abstract, but because it takes the wrong action at the wrong stage.

The tool side $\mathcal{T}$ is a set of reusable graph-algorithm procedures. Each tool $t\in\mathcal{T}$ can be written as $t=(f_t,m_t)$, where $f_t$ is the executable program body and $m_t$ is metadata describing the tool's applicability, such as supported task families, structural regimes, constraint semantics, complexity class, or exactness requirements. Concretely, a reusable tool is characterized not only by its solver logic, but also by its callable API, supported constraint regime, and exactness semantics. These metadata are part of the tool object itself rather than auxiliary annotations, because reusability depends not only on what a tool computes, but also on the conditions under which that computation is compatible with the task.

The retained learning object is not an instruction policy or a tool in isolation, but a \emph{policy pair} $P=(\Pi,\mathcal{T}_{\mathrm{view}})$, where $\mathcal{T}_{\mathrm{view}}\subseteq\mathcal{T}$ denotes the currently visible or reusable tool view. This representation lets us evaluate and retain the instruction policy together with the executable tools it can actually access.

\subsection{Verifier-Centric Solve Process}
Given a task statement $x$, the agent performs a verifier-centric solve chain:
\[
\hat z=\mathrm{Parse}(x;\Pi_{\mathrm{parse}}), \qquad
\mathcal{R}=\mathrm{Retrieve}(x,\hat z,\mathcal{T}_{\mathrm{view}};\Pi_{\mathrm{retrieve}}), \qquad
a_t\sim \Pi_{\mathrm{protocol}}(s_t),
\]
\[
\hat y=\mathrm{Exec}(\hat z,a_t;\Pi_{\mathrm{execute}},\mathcal{R}), \qquad
V=\mathrm{Verify}(\hat y,y^\ast;\hat z,z^\ast).
\]
Here the execution side constructs executable inputs from $\hat z$, the current action, and the retrieved assets, while the external verifier judges structured correctness rather than textual surface plausibility.

In EGL-SCA, the verifier returns structured evidence rather than a single scalar. It may expose schema validity, feasibility, exactness, hidden-test behavior, and execution-time error messages. The verifier is therefore not merely a scorer; it is the central diagnostic source of the learning loop. It determines not only whether the current attempt succeeds, but also what kind of corrective update should follow.

\subsection{Structural Credit Assignment}
The core question of EGL-SCA is: given a failed attempt, which component should be updated? We answer this question through \emph{structural credit assignment}. Given a trajectory $\Gamma$ and verifier evidence $V$, the framework first extracts a diagnostic result $D_{\mathrm{diag}}=\mathrm{Diag}(\Gamma,V)$, and then computes a structured attribution object
\[
C=\mathcal{C}(\Gamma,V,D_{\mathrm{diag}}).
\]
At a coarse level, each unsuccessful attempt is routed to one of four update targets:
$\mathrm{Route}(\Gamma,V)\in\{\texttt{instruction},\texttt{tool\_selection},\texttt{tool\_logic},\texttt{no\_op}\}$.

To define this routing rule, we distinguish three broad error sources: \emph{reconstruction error}, \emph{selection error}, and \emph{tool-logic (algorithmic residual) error}. These correspond to three simple questions: Was the task parsed incorrectly? Was the wrong tool chosen? Or did the chosen tool still fail even though the task and tool match were already reasonable? Concretely, we write $\mathcal{L}_{\mathrm{parse}}=d(\hat z,z^\ast)$, $\mathcal{L}_{\mathrm{sel}}=\Delta_{\mathrm{compat}}(t_{\mathrm{sel}},\hat z)$, and $\mathcal{L}_{\mathrm{alg}}=\Delta_{\mathrm{verify}}(\hat y,y^\ast)$, where $d(\cdot,\cdot)$ is a structured discrepancy measure, $\Delta_{\mathrm{compat}}$ measures tool-task incompatibility, and $\Delta_{\mathrm{verify}}$ denotes the residual verifier-level failure. The routing logic is simple: EGL-SCA blames the earliest stage that clearly failed. If parsing failed, we update the instruction side. Otherwise, if the chosen tool is incompatible with the recovered structure, we update tool selection. Otherwise, if the tool is compatible but still fails, we route the failure to tool-logic repair. If none of these conditions applies, we make no update.
This compact form is a conceptual summary. In the actual implementation, these conditions are instantiated by thresholded diagnostics derived from verifier evidence and trajectory-level signals.

This top-level route is still insufficient for graph reasoning, because many instruction-side failures occur at different stages of the solve process. A failure may be caused by missing required structural fields, an incorrect retrieval-or-reuse transition, a missing protocol escalation, a premature answer before execution, a malformed task input, or a failure to run a candidate tool. EGL-SCA therefore does not merely conclude that the instruction side failed; it further asks \emph{which instruction-side sub-process first deviated from a valid solve path}. This finer localization is essential because a parsing deficiency and an execution-formatting deficiency should not trigger the same update.

The update step is equally simple: once $C$ identifies the failing stage, the framework dispatches to the matching repair operator,
\[
\mathcal{U}(A;C)=\mathcal{U}_{k}(A;C), \qquad
k\in\{\mathrm{parse},\mathrm{retrieve},\mathrm{protocol},\mathrm{execute},\mathcal{T},\mathrm{id}\},
\]
where $k$ denotes reconstruction repair, retrieval/selection repair, transition repair, executable-format repair, tool-side repair, or no update, respectively.
This conditional structure is the key distinction between EGL-SCA and undifferentiated reflection. We do not optimize by simply producing a better post hoc summary; instead, verifier evidence is routed to the correct update operator according to the inferred failure source.

\subsection{Typed Task Interface}
Graph tasks differ not only in graph size, but also in what counts as a complete specification and what output semantics count as correct. To capture this structure, we define for each task family $f$ a typed task interface $\mathcal{I}_f=(\mathcal{S}^{\mathrm{req}}_f,\mathcal{S}^{\mathrm{opt}}_f,E_f)$, where $\mathcal{S}^{\mathrm{req}}_f$ is the set of required fields, $\mathcal{S}^{\mathrm{opt}}_f$ is the set of optional fields, and $E_f$ describes the exactness semantics of the family.

This interface is not an auxiliary annotation layer; it is a structural prior for credit assignment. It makes missing structural fields diagnosable, enables family-specific reflective updates, and explicitly distinguishes semantic regimes such as exact, feasible, and approximate outputs. In this sense, the typed task interface is not separate from structural credit assignment, but one of the structural foundations that allows it to operate at higher resolution. See Appendix~\ref{app:typed_interfaces} for the task-family interface table.

\subsection{Instruction Evolution and Tool-Side Growth/Repair}
When structural credit assignment routes a failure to the instruction side, EGL-SCA performs reflective instruction mutation. Let $C$ denote the current attribution object. A child policy is generated as $\Pi'\sim \mathcal{M}_{\Pi}(\Pi;\Gamma,V,C,\mathcal{I}_f)$, where $\mathcal{M}_{\Pi}$ is a mutation operator conditioned on the trajectory, verifier evidence, attribution result, and family-specific interface. Its purpose is not to rewrite the entire policy indiscriminately, but to produce localized, source-aligned, and verifiable repairs.

When failure is routed to the tool side, EGL-SCA triggers tool growth or repair depending on the circumstance. If the current task lacks a compatible capability, the system may propose a new candidate tool from the current tool space; if a compatible tool exists but behaves incorrectly, the system performs tool-side repair or variation. Abstractly, we write
\[
t'\sim \mathcal{M}_{\mathcal{T}}(\mathcal{T};\Gamma,V,C).
\]
Unlike unconstrained code generation, any candidate tool must satisfy verifier-based acceptance before entering the reusable toolbox. This rule ensures that tool growth is grounded in observed task performance rather than drifting into an unstructured repository of unverifiable programs.

\subsection{Pareto-Style Policy-Pair Selection}
Because graph reasoning quality cannot be summarized adequately by a single scalar, EGL-SCA retains policy pairs through multi-objective selection. For each policy pair $P$, we compute an objective vector $\mathbf{F}(P)=\big[S(P),G(P),R(P),Q(P)\big]$, where $S(P)$ is validation success, $G(P)$ is generality across evaluation buckets, $R(P)$ is protocol reliability, and $Q(P)$ is a parsimony measure favoring concise, stable, and low-redundancy behavior. Rather than collapsing these criteria into one score, we retain the non-dominated frontier
\[
\mathcal{F}=\{P\mid \nexists P'\text{ such that }\mathbf{F}(P')\succ \mathbf{F}(P)\},
\]
where $\succ$ denotes Pareto dominance. This rule allows training to preserve policy pairs that are not yet best in raw task success but are already stronger in stability or generality. Final benchmark evaluation is then performed with a frozen pair selected from this frontier.

Taken together, these components define EGL-SCA as a verifier-centric framework for dual-space self-improvement. The instruction side learns to reconstruct and manage graph tasks more reliably, the tool side grows or repairs executable graph procedures when necessary, and structural credit assignment determines how failure evidence should flow between them. The resulting system is neither prompt optimization alone nor program search in isolation, but a coordinated learning loop over language policies and executable graph assets. See Appendix~\ref{app:algorithm_sca} for the training-loop pseudocode and route table.

\section{Experiments}

\begin{table}[!t]
    \centering
    \small
    \renewcommand{\arraystretch}{1.08}
    \caption{External performance and training-time mechanism statistics for EGL-SCA and its baselines.}
    \label{tab:exp_tables}
    
    \definecolor{winGreen}{HTML}{1A855A}
    \definecolor{lossRed}{HTML}{C1272D}
    
    \newcommand{\win}[1]{$_{\!\color{winGreen}{+#1}}$}
    \newcommand{\loss}[1]{$_{\!\color{lossRed}{-#1}}$}
    
    \begin{subtable}{\textwidth}
        \centering
        \caption{External benchmark success rates. Best in \textbf{bold}, second \underline{underlined}. Subscripts denote performance deltas: ablations versus the Full model, and the Full model versus the strongest baseline (AgentSquare).}
        \label{tab:main_results}
        \begin{threeparttable}
        \begin{tabular*}{\textwidth}{@{\extracolsep{\fill}} l *{5}{c@{}l} @{}}
            \toprule
            \textbf{Method} & 
            \multicolumn{2}{c}{\textbf{GraphInstruct}} & 
            \multicolumn{2}{c}{\textbf{GraphArena}} & 
            \multicolumn{2}{c}{\textbf{NLGraph}} & 
            \multicolumn{2}{c}{\textbf{G-Real}} & 
            \multicolumn{2}{c}{\textbf{Avg}} \\
            \midrule
            \multicolumn{11}{@{}l}{\textit{\textbf{Prompting \& Self-Improvement Baselines (No Tools)}}} \\
            \midrule
            Direct Prompting       & 0.525 & & 0.160 & & 0.245 & & 0.015 & & 0.236 & \\
            Chain-of-Thought (CoT) & 0.645 & & 0.420 & & 0.300 & & 0.020 & & 0.346 & \\
            Few-Shot               & 0.400 & & 0.080 & & 0.150 & & 0.065 & & 0.174 & \\
            Reflexion              & 0.175 & & 0.045 & & 0.170 & & 0.055 & & 0.111 & \\
            ExpeL                  & 0.380 & & 0.180 & & 0.210 & & 0.055 & & 0.206 & \\
            GEPA                   & 0.640 & & 0.475 & & 0.215 & & 0.000 & & 0.333 & \\
            \midrule
            \multicolumn{11}{@{}l}{\textit{\textbf{Tool-Augmented Agents (Fixed Toolbox)}}} \\
            \midrule
            AgentSquare            & \underline{0.875} & & \textbf{1.000} & & \underline{0.635} & & \textbf{1.000} & & \underline{0.878} & \\
            ReAct                  & 0.405 & & 0.655 & & 0.590 & & 0.195 & & 0.461 & \\
            MA-GTS\tnote{\dag}     & \multicolumn{2}{c}{--} & \textbf{1.000} & & 0.629 & & 0.720 & & 0.783 & \\
            FixedSolver            & 0.575 & & 0.670 & & 0.580 & & \underline{0.795} & & 0.655 & \\
            \midrule
            \multicolumn{11}{@{}l}{\textit{\textbf{EGL-SCA (Ours) \& Ablations}}} \\
            \midrule
            w/o SCA Routing          & 0.875 & \loss{0.045} & 0.995 & & 0.930 & \loss{0.040} & 0.780 & \loss{0.015} & 0.895 & \loss{0.025} \\
            w/o Tool Use             & 0.685 & \loss{0.235} & 0.360 & \loss{0.635} & 0.445 & \loss{0.525} & 0.030 & \loss{0.765} & 0.380 & \loss{0.540} \\
            w/o Tool Growth          & 0.885 & \loss{0.035} & 0.995 & & 0.930 & \loss{0.040} & 0.695 & \loss{0.100} & 0.876 & \loss{0.044} \\
            w/o Instruction Evol.    & 0.885 & \loss{0.035} & 0.910 & \loss{0.085} & 0.885 & \loss{0.085} & 0.760 & \loss{0.035} & 0.860 & \loss{0.060} \\
            w/o Protocol-Reliability Obj & 0.890 & \loss{0.030} & 0.995 & & 0.945 & \loss{0.025} & 0.795 & & 0.906 & \loss{0.014} \\
            \addlinespace 
            \textbf{EGL-SCA (Full)}  & \textbf{0.920} & \win{0.045} & \underline{0.995} & \loss{0.005} & \textbf{0.970} & \win{0.335} & \underline{0.795} & \loss{0.205} & \textbf{0.920} & \win{0.042} \\
            \bottomrule
        \end{tabular*}
        \begin{tablenotes}
            \small
            \item[\dag] MA-GTS failed to produce valid results on GraphInstruct due to toolbox limitations.
        \end{tablenotes}
        \end{threeparttable}
    \end{subtable}

    \vspace{0.55em} 

    \begin{subtable}{\textwidth}
        \centering
        \caption{Training-time mechanism statistics.}
        \label{tab:mechanism_stats}
        \begin{tabular*}{\textwidth}{@{\extracolsep{\fill}} l ccccc @{}}
            \toprule
            \multirow{2}{*}{\textbf{Method}} & \textbf{Protocol} & \textbf{Accepted Inst.} & \textbf{Packaged} & \textbf{Used} & \textbf{Frontier} \\
            & \textbf{Reliability} & \textbf{Mutations} & \textbf{Tools} & \textbf{Tools} & \textbf{Size} \\
            \midrule
            w/o SCA Routing          & 0.788 & 22 & 39 & 32 & 4 \\
            w/o Tool Use             & 0.775 & 10 & 0  & 0  & 4 \\
            w/o Instruction Evol.    & 0.817 & 0  & 40 & 26 & 2 \\
            \addlinespace
            \textbf{EGL-SCA (Full)}  & \textbf{0.936} & 2  & 27 & 26 & 3 \\
            \bottomrule
        \end{tabular*}
    \end{subtable}
\end{table}

\subsection{Experimental goals and setup}

We design our empirical evaluation to systematically answer four core research questions (\textbf{RQs}):
\begin{itemize}[leftmargin=*, nosep]
    \item $\mathbf{RQ_1}$ \textbf{(Overall Superiority):} Does EGL-SCA consistently outperform pure-prompting methods and fixed-toolbox baselines on natural-language graph reasoning benchmarks?
    \item $\mathbf{RQ_2}$ \textbf{(Mechanism Ablation):} Is this performance gain driven by the synergistic effect of the dual-space loop (i.e., structured credit assignment, instruction evolution, and tool growth) rather than any isolated component?
    \item $\mathbf{RQ_3}$ \textbf{(Task-Family Generalization):} Does our framework produce balanced, structural capability gains across diverse graph task families, rather than merely inflating an aggregate score?
    \item $\mathbf{RQ_4}$ \textbf{(Curriculum Dynamics):} How does the training distribution affect the emergence of dual-space capabilities, particularly when comparing a progressive $1 \!\rightarrow\! 4$ curriculum against a fixed-D4 setting?
\end{itemize}

Unless otherwise stated, all reported training and benchmark runs use \texttt{gpt-5.4-nano} as the base LLM, with the same verifier-centric interaction protocol across methods.

The internal training side uses a unified graph-task generator covering 19 task families (e.g., shortest path, TSP, topological sorting, bipartite matching). The generator varies graph scale, structural regime, active constraints, and exactness semantics, exposing heterogeneous failures in parsing, protocol behavior, tool selection, and tool-logic errors. Each generated instance is assigned one of four difficulty levels, D1--D4, which act as a composite control over graph size, constraint complexity, and verifier strictness. Here, \emph{fixed-D4} denotes training entirely at the hardest level, whereas \emph{progressive $1 \!\rightarrow\! 4$} denotes training that starts from D1 and gradually advances to D4. See Appendix~\ref{app:typed_interfaces} for the grouped task-interface specification.

\paragraph{Dynamic data synthesis and curriculum.}
Training instances are dynamically synthesized on-the-fly rather than sampled statically. Each episode selects a task family via deterministic round-robin across 19 categories. A generator instantiates a canonical graph, queries, and constraints based on the current difficulty and seed, which a reference algorithm solves to yield ground-truth data strictly for verifier evaluation. A rule-based verbalizer then translates the instance into a natural-language statement. The agent observes only this text; the underlying \texttt{task\_input} and solutions remain hidden. For the pass-based curriculum, each family's difficulty independently advances from D1 to D4 after two successful solves at its current tier.

On the external side, we evaluate on four standardized benchmarks: \textbf{GraphInstruct, GraphArena, NLGraph, and G-Real}~\citep{luo2024graphinstruct,tang2024grapharena,wang2023nlgraph,yuan2025magts}. All main results use a verifier-centric protocol, where success is defined by structured correctness rather than surface-form similarity. See Appendix~\ref{app:benchmark_protocol} for benchmark and baseline protocol details, Appendix~\ref{app:reproducibility} for implementation details, and Appendix~\ref{app:training_cost} for training-cost accounting.

\subsection{Main Results: The Necessity of Dual-Space Co-Evolution}

Table~\ref{tab:exp_tables}\subref{tab:main_results} compares EGL-SCA against various baselines under a unified external evaluation protocol. EGL-SCA (Full) achieves a state-of-the-art average success rate of \textbf{0.920}, significantly outperforming the strongest fixed-toolbox workflow (AgentSquare, 0.878) and graph-native pipelines~\citep{shang2024agentsquare,yuan2025magts}. 

The collapse of prompting baselines (e.g., CoT at 0.346, Reflexion at 0.111) reveals a fundamental misalignment: \textbf{LLMs often mask structural hallucinations with linguistic fluency.} In exact graph reasoning, a single missed constraint invalidates the entire trajectory. Language-only reflection treats this as a text-generation error, repeatedly tweaking the prompt without addressing the underlying algorithmic deficit. This proves that externalizing computation to a tool space is not optional, but mandatory.

Furthermore, the ablations confirm the necessity of the dual-space loop. Removing tool use causes the most severe degradation (dropping to 0.380). However, removing instruction evolution also reduces the average to 0.860, demonstrating that tools alone cannot solve the natural-language-to-executable-structure binding problem. We further provide a three-seed sensitivity analysis in Appendix~\ref{app:seed_sensitivity}.

\subsection{Mechanism Diagnostics: SCA Prevents Update Thrashing}
  \begin{figure*}[!t]
    \centering
    \begin{subfigure}[t]{0.49\textwidth}
      \centering
      \includegraphics[width=\linewidth]{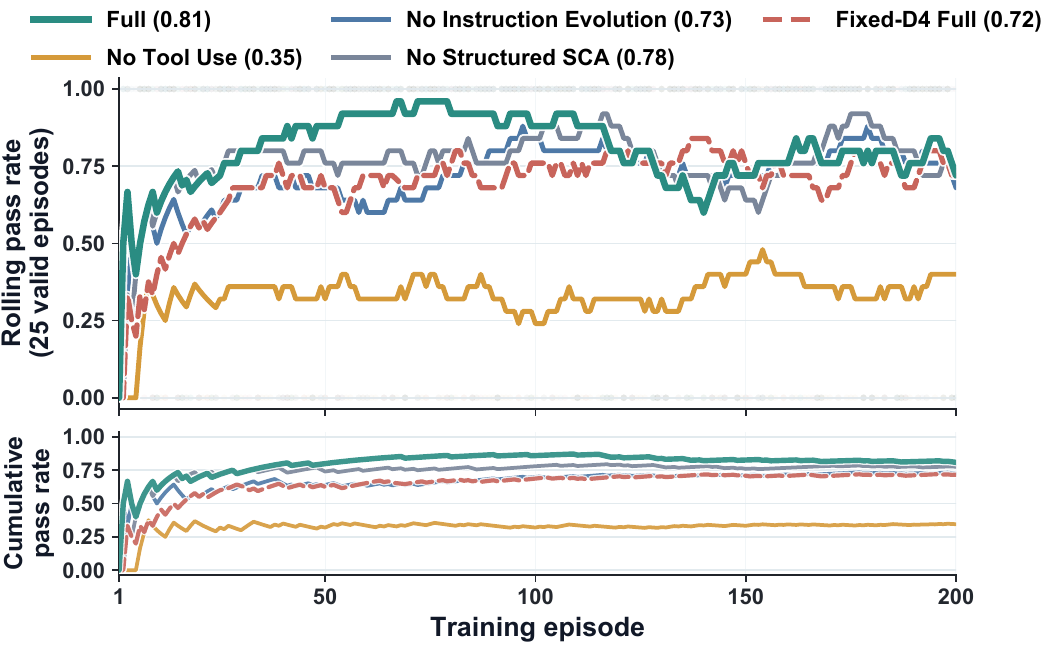}
      \caption{EGL-SCA learns faster and saturates higher than its ablations.}
      \label{fig:training_curves}
    \end{subfigure}\hfill
    \begin{subfigure}[t]{0.49\textwidth}
      \centering
      \includegraphics[width=\linewidth]{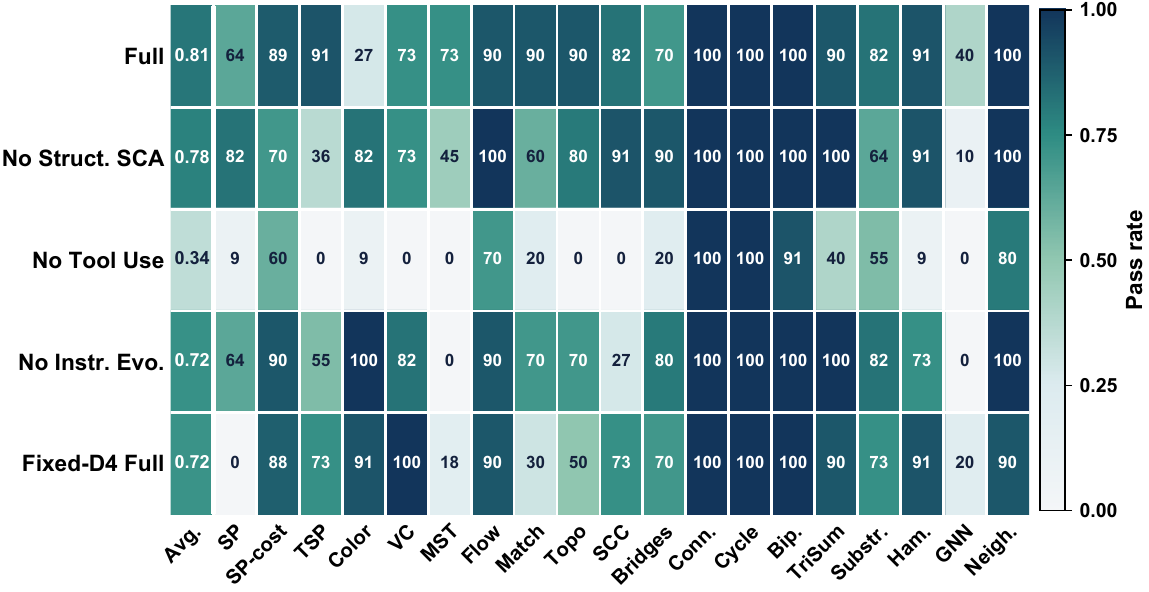}
      \caption{The full method remains more balanced across graph families.}
      \label{fig:task_family_heatmap}
    \end{subfigure}

    \vspace{0.15em}

    \begin{subfigure}[t]{0.49\textwidth}
      \centering
      \includegraphics[width=\linewidth]{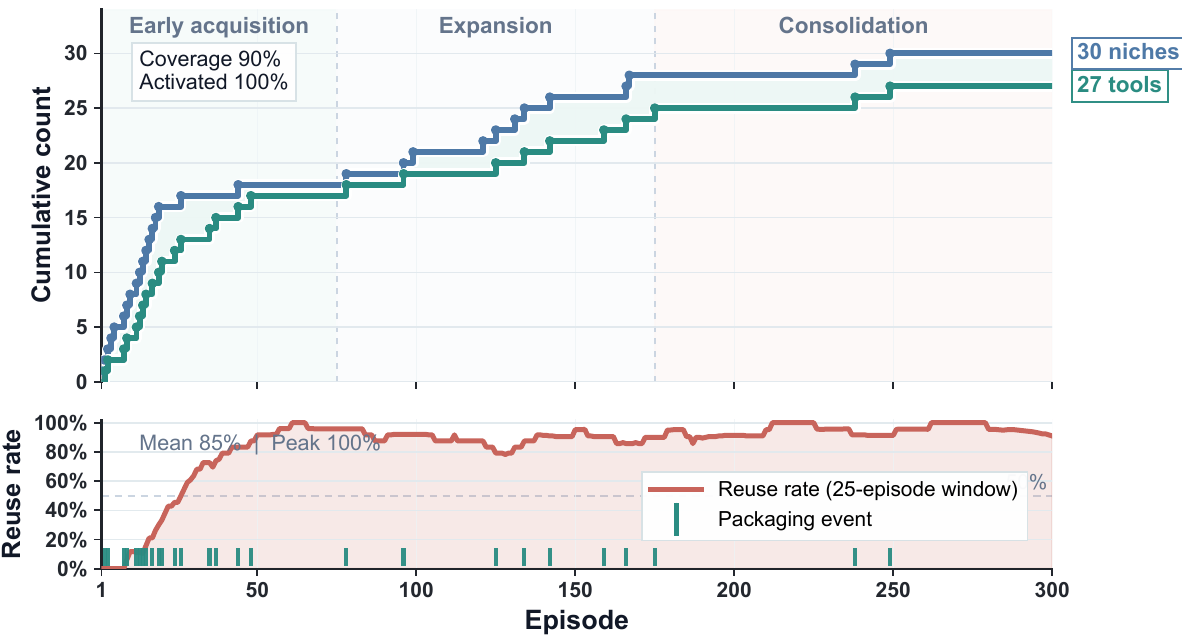}
      \caption{Tool discovery quickly turns into stable reuse.}
      \label{fig:tool_growth}
    \end{subfigure}\hfill
    \begin{subfigure}[t]{0.49\textwidth}
      \centering
      \includegraphics[width=\linewidth]{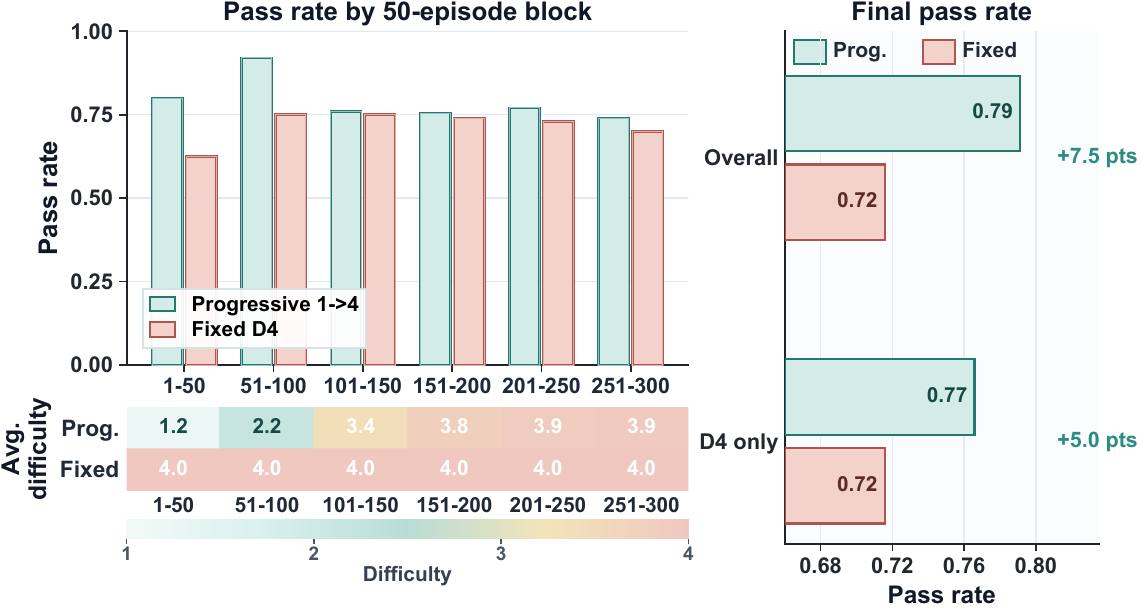}
      \caption{Progressive 1$\rightarrow$4 training beats fixed-D4 even on D4-only tasks.}
      \label{fig:curriculum_fixed}
    \end{subfigure}
    \caption{Internal diagnostics for EGL-SCA: learning dynamics, family-level structure, tool reuse, and curriculum effects.}
    \label{fig:exp_panels}
  \end{figure*}

Table~\ref{tab:exp_tables}\subref{tab:mechanism_stats} reports training-time mechanism statistics, revealing exactly \emph{how} EGL-SCA learns. Without SCA Routing, the system exhibits a crucial failure mode we term \textbf{``update thrashing''}. The agent acts like a blind optimizer: when a tool-logic error occurs, it mistakenly punishes the parsing instruction (accepting 22 redundant mutations) and hallucinates 39 overlapping tools. This misattribution degrades protocol reliability (0.788) because the agent constantly unlearns previously successful behaviors.

In contrast, EGL-SCA (Full) demonstrates that \textbf{learning efficiency comes from precise blame assignment, not update frequency}. By routing failures accurately, the full system maintains an exceptional protocol reliability of 0.936 while accepting only 2 instruction mutations and packaging 27 highly reusable tools. It forms a compact, parsimonious combination of instruction behavior and toolbox contents. See Appendices~\ref{sec:case_studies} and~\ref{sec:prompt_evolution} for trace-level and prompt-evolution examples.

\subsection{Training Dynamics: Stable Convergence via Structured Routing}

Figure~\ref{fig:exp_panels}\subref{fig:training_curves} shows the internal learning dynamics. The full method rises quickly and reaches approximately 0.809 cumulative pass rate by episode 200. In contrast, \emph{w/o Tool Use} stagnates between 0.25 and 0.45, reaffirming that generative reasoning alone hits a hard ceiling. Interestingly, \emph{w/o SCA Routing} produces early local gains but remains volatile and less efficient overall, proving that unstructured updates eventually plateau due to accumulated contradictory edits.

\subsection{Task-Level Analysis: Where and Why Generative Reasoning Fails}

Figure~\ref{fig:exp_panels}\subref{fig:task_family_heatmap} decomposes performance across the 19 task families. The full method remains highly balanced (internal average $\sim$0.81). As expected, \emph{w/o Tool Use} nearly fails completely on exact combinatorial tasks such as TSP, topological sorting, and MST, which require stable graph traversal and decomposition routines. 

The degradation patterns of the ablations are highly diagnostic. \emph{w/o Instruction Evolution} loses substantially on MST and bipartite matching, indicating these tasks heavily rely on language-side adaptation for reuse decisions and valid execution payloads. \emph{w/o SCA Routing} degrades severely when constraint parsing and tool compatibility heavily interact, proving that structured attribution is crucial when resolving multi-step dependencies.

\subsection{Toolbox Growth and Reuse Dynamics}

Figure~\ref{fig:exp_panels}\subref{fig:tool_growth} analyzes toolbox growth. The system observes 30 tool niches and packages 27 tools. More importantly, the activation rate of these tools reaches 100\%, and the rolling reuse rate peaks at 100\% (mean $\sim$85\%). This temporal dimension illustrates that EGL-SCA first covers frequent niches, consolidates tools during expansion, and rapidly enters a \textbf{reuse-dominated regime}. The tool side acts as a verifier-guided repository of reusable algorithmic tools rather than a static, bloated toolbox; Appendix~\ref{app:frozen_tool_inventory} lists the frozen toolbox used for benchmark evaluation.

\subsection{Curriculum Dynamics: Bootstrapping Credit Assignment}

Figure~\ref{fig:exp_panels}\subref{fig:curriculum_fixed} reveals that progressive $1 \!\rightarrow\! 4$ training outpaces the fixed-D4 setting (0.791 vs. 0.716 overall pass rate). Even when restricted purely to D4 evaluation samples, progressive training maintains a 5.0-point advantage. 

Why does curriculum training yield stronger final capabilities? In high-difficulty graphs, the trajectory is extremely long and errors compound rapidly, creating a \textbf{dense credit assignment bottleneck}. The verifier's failure signal becomes too noisy to localize. Progressive training allows the agent to first calibrate its parsing and protocol reliability ($\Pi$ space) on simpler graphs. Once the instruction policy is stabilized, the system can reliably isolate and target complex tool-logic errors ($\mathcal{T}$ space) without interference from basic parsing failures.

\section{Conclusion}

We presented EGL-SCA, a verifier-centric framework for graph reasoning agents that co-evolves instruction policies and executable tools through structural credit assignment. The core idea is simple: when a graph reasoning attempt fails, the agent should not blindly rewrite everything, but should identify whether the error came from parsing, protocol behavior, tool selection, or tool-logic errors. Across external benchmarks and internal diagnostics, EGL-SCA shows that this targeted routing leads to stronger accuracy, more reliable execution, and a compact reusable toolbox. These results suggest that future self-improving agents should treat instruction policies and executable programs as coupled learning objects, with verifier evidence deciding where adaptation should occur.

{\small
\bibliographystyle{unsrtnat}
\bibliography{refs}
}

\appendix

\section{Reproducibility and Implementation Details}
\label{app:reproducibility}

\paragraph{Model and decoding.}
All reported benchmark results use the same underlying GPT-5.4-nano model family and the same verifier-centric interaction protocol across methods. Decoding configuration, retry policy, timeout settings, and prompt templates are kept fixed within each comparison group; the exact runnable configuration is recorded with the anonymous experiment artifacts. This design is intended to make differences in Table~\ref{tab:main_results} attributable to the agent protocol and update mechanism rather than to per-method decoding changes.

\paragraph{Training budget and randomness.}
The main training runs use a 200-episode budget unless otherwise stated. The curriculum comparison uses 300 episodes to compare progressive $1\!\rightarrow\!4$ training with fixed-D4 training under the same high-level protocol. All main runs are generated with seed \texttt{20260413}. Randomness enters through task sampling, LLM generation, candidate-tool mutation, and policy-pair selection, and all logged summaries retain the seed and run identifier.

\paragraph{Execution and verification.}
Executable graph tools are run in a Python execution environment and judged by task-family verifiers. The verifier checks structured correctness, including schema validity, feasibility, exactness, hidden-test behavior, and constraint satisfaction, rather than textual similarity. Success rate is computed as the number of verifier-passed tasks divided by the number of evaluated tasks.

\paragraph{Logged evidence.}
Each solve attempt records the task family, difficulty, pass/fail result, selected tool if any, whether a candidate tool ran, verifier messages, and structured credit-assignment diagnostics. The most important diagnostic fields include \texttt{task\_kind}, \texttt{difficulty}, \texttt{passed}, \texttt{tool\_used}, \texttt{candidate\_ran}, \texttt{credit\_assignment.route}, \texttt{typed\_missing\_slots}, and \texttt{protocol\_signature}. These fields support the mechanism statistics in Table~\ref{tab:mechanism_stats} and the internal diagnostics in Figure~\ref{fig:exp_panels}.

Table~\ref{tab:run_eval_config} summarizes the exact configuration used for the main training run and for the frozen benchmark evaluations. Table~\ref{tab:curriculum_details} further details how the pass-based curriculum was instantiated and how much exposure the run allocated to each difficulty band.

\begin{table}[!htbp]
  \centering
  \footnotesize
  \setlength{\tabcolsep}{3pt}
  \renewcommand{\arraystretch}{1.1}
  \caption{Unified training and benchmark-evaluation configuration for the reported full EGL-SCA system.}
  \label{tab:run_eval_config}
  \begin{tabularx}{\textwidth}{@{} >{\raggedright\arraybackslash}p{0.23\textwidth} >{\raggedright\arraybackslash}X >{\raggedright\arraybackslash}X @{}}
    \toprule
    \textbf{Setting} & \textbf{Main Training Run} & \textbf{External Benchmark Evaluation} \\
    \midrule
    Base model & \texttt{gpt-5.4-nano} & \texttt{gpt-5.4-nano} \\
    Student prompt profile & \texttt{weak} & \texttt{weak} frozen instruction genome \\
    Budget & 300 planned episodes; 297 counted after 3 exclusions & 200 cases per benchmark \\
    Attempts & 2 attempts per training episode & 1 attempt per benchmark case \\
    Toolbox policy & fresh toolbox with dynamic growth and repair enabled & frozen reusable toolbox from the selected policy pair \\
    Tool proposal at test time & enabled during training & disabled (\texttt{--no-propose-tool}) \\
    Selection rule & Pareto over success, generality, protocol reliability, and parsimony; re-evaluated every 38 episodes & frozen pair selected under the same objective vector \\
    Population / concurrency & 4 policy pairs; at most 4 active pairs & single frozen pair \\
    Validation / probes & 2 validation cases per kind; 2 tool-probe cases & none; direct benchmark evaluation \\
    Decoding & temperature 0.0 & temperature 0.0 \\
    Timeout / retries & 420 s / 12 retries & 300 s / 12 retries \\
    Seed & \texttt{20260413} & \texttt{20260413} \\
    \bottomrule
  \end{tabularx}
\end{table}

\begin{table}[!htbp]
  \centering
  \footnotesize
  \setlength{\tabcolsep}{4pt}
  \renewcommand{\arraystretch}{1.08}
  \caption{Pass-based curriculum configuration and realized difficulty exposure in the main 300-episode run. Advancement was enabled per task family, starting from D1 and promoting after 2 passes at the current difficulty.}
  \label{tab:curriculum_details}
  \begin{tabularx}{\textwidth}{@{} l c c X @{}}
    \toprule
    \textbf{Difficulty} & \textbf{First Seen Episode} & \textbf{Counted Episodes} & \textbf{Interpretation} \\
    \midrule
    D1 & 1   & 52  & Initial easy-stage exposure used to stabilize parsing, protocol discipline, and early tool discovery. \\
    D2 & 40  & 40  & First promoted band after per-task pass thresholds began to trigger. \\
    D3 & 78  & 47  & Intermediate band where retrieval and compatibility decisions became more consequential. \\
    D4 & 116 & 158 & Longest training phase; all 19 task kinds reached D4 and no task kind remained unfinished. \\
    \bottomrule
  \end{tabularx}
\end{table}

\FloatBarrier

\section{Task Families and Typed Interfaces}
\label{app:typed_interfaces}

The typed task interface specifies which fields must be recovered from natural language, which optional constraints may be active, and what output semantics the verifier expects. Table~\ref{tab:typed_interfaces} summarizes the internal task families used for training and diagnostics. The interface is used to detect missing structural slots, distinguish exact versus feasible outputs, and localize failures such as malformed payloads or output-schema mismatches.

\begin{table}[!htbp]
  \centering
  \footnotesize 
  \setlength{\tabcolsep}{2.5pt}
  \renewcommand{\arraystretch}{1.16} 
  \caption{Typed task interfaces used by the internal graph-task generator. The interface specifies the exact structured payload requirements and evaluation semantics for each task family.}
  \label{tab:typed_interfaces}
  
  \begin{tabularx}{\textwidth}{@{} 
    >{\hsize=0.85\hsize\raggedright\arraybackslash}X 
    >{\hsize=0.90\hsize\raggedright\arraybackslash}X 
    >{\hsize=1.35\hsize\raggedright\arraybackslash}X 
    >{\hsize=0.90\hsize\raggedright\arraybackslash}X 
    >{\hsize=1.00\hsize\raggedright\arraybackslash}X @{}}
    \toprule
    \textbf{Task Family} & \textbf{Representative Tasks} & \textbf{Required Fields} & \textbf{Output Semantics} & \textbf{Typical Failure Signal} \\
    \midrule
    
    \textbf{Path Queries} 
    & Shortest path, \newline SP cost 
    & \texttt{graph.nodes}, \texttt{graph.edges}, \texttt{query.source}, \texttt{query.target}, opt. weights 
    & Exact path or exact cost 
    & Missing endpoint, invalid path, wrong distance \\
    \addlinespace
    
    \textbf{Routing \& \newline Coverage} 
    & TSP, Hamiltonian path 
    & \texttt{graph.nodes}, \texttt{graph.edges}, \texttt{graph.directed}, opt. weights 
    & Exact tour/path or infeasibility cert. 
    & Repeated node, non-edge step, suboptimal cost \\
    \addlinespace
    
    \textbf{Labeling \& \newline Covering} 
    & Coloring, vertex cover 
    & \texttt{graph.nodes}, \texttt{graph.edges}, max-color constraint or cover budget 
    & Exact small-case optimum or feasible assign. 
    & Uncovered edge, invalid color assignment \\
    \addlinespace
    
    \textbf{Optimization \newline Flow} 
    & MST, max flow, bipartite matching 
    & Weighted/capacitated edges, source/sink or bipartition (if required) 
    & Exact optimum value and witness 
    & Disconnected tree, capacity violation, unmatched edge \\
    \addlinespace
    
    \textbf{Decomp. \& \newline Order} 
    & Topological sort, SCC, bridges 
    & \texttt{graph.nodes}, \texttt{graph.edges}, \texttt{graph.directed} 
    & Exact ordering, components, or bridge set 
    & Cycle in order, wrong component partition \\
    \addlinespace
    
    \textbf{Connectivity \newline Checks} 
    & Connectivity, cycle detection, bipartite checking 
    & \texttt{graph.nodes}, \texttt{graph.edges}, \texttt{graph.directed} (when active) 
    & Boolean answer with witness if needed 
    & False witness, missing component, parity conflict \\
    \addlinespace
    
    \textbf{Local \newline Substructure} 
    & Triangle max sum, common neighbors 
    & Node labels or weights, query node pair (if required) 
    & Exact set, count, or weighted score 
    & Wrong node set, missed maximum triangle \\
    \addlinespace
    
    \textbf{Pattern \newline Matching} 
    & Substructure \newline matching 
    & Host graph, pattern graph, matching constraints 
    & Feasible or exact mapping 
    & Non-injective map, unmatched pattern edge \\
    \addlinespace
    
    \textbf{Message \newline Passing} 
    & GNN-style \newline message passing 
    & Graph, initial node states, aggregation rule, number of rounds 
    & Exact final states 
    & Wrong aggregation, wrong round count \\
    
    \bottomrule
  \end{tabularx}
\end{table}
\FloatBarrier

\section{External Benchmarks and Baseline Protocol}
\label{app:benchmark_protocol}

All external datasets~\citep{luo2024graphinstruct,tang2024grapharena,wang2023nlgraph,yuan2025magts} are converted to a common task format before evaluation. The agent observes a natural-language problem statement, while the verifier uses the canonical structured object and reference answer to judge success. Table~\ref{tab:external_benchmarks} summarizes the four external benchmarks, and Table~\ref{tab:baseline_protocols} summarizes the baseline protocols used in Table~\ref{tab:main_results}.

\begin{table}[!htbp]
  \centering
  \small
  \setlength{\tabcolsep}{3pt}
  \renewcommand{\arraystretch}{1.06}
  \caption{External benchmark protocol.}
  \label{tab:external_benchmarks}
  \begin{tabularx}{\textwidth}{@{} l >{\raggedright\arraybackslash}X >{\raggedright\arraybackslash}X >{\raggedright\arraybackslash}X @{}}
    \toprule
    \textbf{Benchmark} & \textbf{Input form} & \textbf{Evaluation target} & \textbf{Main stress} \\
    \midrule
    GraphInstruct & natural-language graph instructions & structured graph answer & graph understanding and schema recovery \\
    GraphArena & graph computation problems & algorithmic answer or witness & exact graph computation \\
    NLGraph & natural-language graph tasks & parsed graph reasoning answer & language-to-graph reconstruction \\
    G-Real & real-world style graph problems & verifiable graph solution & realistic constraints and tool coverage \\
    \bottomrule
  \end{tabularx}
\end{table}

\begin{table}[!htbp]
  \centering
  \footnotesize
  \setlength{\tabcolsep}{3pt}
  \renewcommand{\arraystretch}{1.08}
  \caption{Benchmark scale statistics for the four external evaluations used in the main paper.}
  \label{tab:benchmark_scale_stats}
  \begin{tabularx}{\textwidth}{@{} l c c >{\raggedright\arraybackslash}X c c @{}}
    \toprule
    \textbf{Benchmark} & \textbf{Cases} & \textbf{Task Kinds} & \textbf{Composition} & \textbf{Avg Nodes} & \textbf{Avg Edges} \\
    \midrule
    GraphInstruct & 200 & 8 & bipartite\_check 25; connectivity 25; cycle 25; hamilton 25; max\_flow 25; shortest\_path\_cost 25; topological\_sort 25; triangle\_max\_sum 25 & 26.04 & 89.30 \\
    NLGraph & 200 & 5 & cycle 68; gnn\_sum 13; hamilton 40; max\_flow 39; shortest\_path 40 & 17.46 & 34.64 \\
    GraphArena & 200 & 2 & common\_neighbors 100; tsp 100 & 8.52 & 24.64 \\
    G-Real & 200 & 3 & coloring 67; tsp 67; vertex\_cover 66 & 18.84 & 111.44 \\
    \bottomrule
  \end{tabularx}
\end{table}

\begin{table}[!htbp]
  \centering
  \footnotesize 
  \setlength{\tabcolsep}{3pt}
  \caption{Comparison of baseline methodologies. Unlike existing methods that either rely purely on language updates or use a fixed tool inventory, EGL-SCA achieves true dual-space co-evolution.}
  \label{tab:baseline_protocols}
  \renewcommand{\arraystretch}{1.08}
  \begin{tabularx}{\textwidth}{@{} l c c >{\raggedright\arraybackslash}X @{}}
    \toprule
    \textbf{Method} & \textbf{Tool Access} & \textbf{Update Mechanism} & \textbf{Key Characteristics} \\
    \midrule
    
    \multicolumn{4}{@{}l}{\textit{Static Prompting (Language-Only)}} \\
    Direct Prompting & None & None & Zero-shot generation under verifier protocol. \\
    Few-Shot         & None & None & Static in-context demonstrations. \\
    Chain-of-Thought & None & None & Step-by-step reasoning traces. \\
    \addlinespace
    
    \multicolumn{4}{@{}l}{\textit{Self-Refinement \& Evolution (Language-Only)}} \\
    Reflexion & None & Verbal Reflection & Iterative prompt correction via verbal feedback. \\
    ExpeL     & None & Experience Pool   & Accumulates textual insights from trajectories. \\
    GEPA      & None & Prompt Evolution  & Evaluator-guided prompt and program refinement. \\
    \addlinespace
    
    \multicolumn{4}{@{}l}{\textit{Tool-Augmented \& Graph-Native Agents (Fixed Assets)}} \\
    ReAct       & Fixed APIs & In-context Action & Interleaves reasoning with external tool calls. \\
    FixedSolver & Fixed Algorithms & None & Routes directly to predefined solver inventory. \\
    AgentSquare & Fixed Workflows  & Modular Search & Strongest fixed-toolbox baseline (SOTA modular). \\
    MA-GTS      & Graph-native APIs& Multi-Agent & Collaborative framework (fails on GraphInstruct). \\
    \midrule
    
    \textbf{EGL-SCA (Ours)} & \textbf{Dynamic Growth} & \textbf{Dual-Space SCA} & \textbf{Structurally routes credit to co-evolve prompts and synthesize executable algorithms.} \\
    \bottomrule
  \end{tabularx}
\end{table}

Prompt-only baselines never access executable tools. ReAct and AgentSquare~\citep{yao2023react,shang2024agentsquare} are evaluated as tool or workflow baselines with a fixed inventory. MA-GTS~\citep{yuan2025magts} is treated as a graph-native multi-agent baseline; its missing GraphInstruct entry in Table~\ref{tab:main_results} reflects an interface and toolbox-coverage limitation rather than a failed score being hidden.

\FloatBarrier

\section{Seed Sensitivity Analysis}
\label{app:seed_sensitivity}

To rigorously assess the robustness of EGL-SCA against variations in initialization and trajectory sampling, we conduct a sensitivity analysis across three independent random seeds: \texttt{20260414}, \texttt{20260415}, and \texttt{20260416}. Each seed instantiates a fresh dual-space learning environment to train the agent autonomously, after which the Pareto-optimal policy pair is frozen for zero-shot evaluation on the standard external benchmarks. 

Given the substantial computational overhead of executing three complete training-and-evaluation lifecycles, this sweep employs a constrained budget of 150 training episodes per seed. Consequently, this analysis acts as a conservative lower-bound estimation of stability rather than a direct substitute for the primary 300-episode evaluations. Under this truncated schedule, several challenging task families do not accumulate sufficient exposure to complete curriculum promotion to the highest difficulty tier (D4).

\begin{table}[!htbp]
  \centering
  \small
  \renewcommand{\arraystretch}{1.2}
  \caption{Seed sensitivity on external benchmarks under a constrained 150-episode training budget. Each benchmark evaluation contains 200 cases per seed. Values denote verifier pass rates (\%).}
  \label{tab:seed_sensitivity_benchmark}
  \begin{tabularx}{\textwidth}{@{} l *{3}{>{\centering\arraybackslash}X} c c @{}}
    \toprule
    \multirow{2}{*}{\textbf{Benchmark}} & \multicolumn{3}{c}{\textbf{Independent Runs}} & \multirow{2}{*}{\textbf{Mean $\pm$ Std.}} & \multirow{2}{*}{\textbf{Pooled Rate}} \\
    \cmidrule(lr){2-4}
    & \textbf{Seed 20260414} & \textbf{Seed 20260415} & \textbf{Seed 20260416} & & \\
    \midrule
    GraphInstruct & 95.5 & 91.5 & 96.0 & 94.3 $\pm$ 2.5 & 566 / 600 = 94.3 \\
    NLGraph       & 95.5 & 99.0 & 95.0 & 96.5 $\pm$ 2.2 & 579 / 600 = 96.5 \\
    GraphArena    & 97.5 & 100.0 & 95.0 & 97.5 $\pm$ 2.5 & 585 / 600 = 97.5 \\
    G-Real        & 79.5 & 77.0 & 86.5 & 81.0 $\pm$ 4.9 & 486 / 600 = 81.0 \\
    \midrule
    \textbf{Overall} & \textbf{92.0} & \textbf{91.9} & \textbf{93.1} & \textbf{92.3 $\pm$ 0.7} & \textbf{2216 / 2400 = 92.3} \\
    \bottomrule
  \end{tabularx}
\end{table}

\begin{table}[!htbp]
  \centering
  \small
  \renewcommand{\arraystretch}{1.2}
  \caption{Pooled case-level confidence intervals for the three-seed sensitivity sweep. The 95\% Wilson score intervals are computed over the aggregated execution traces ($n=600$ per benchmark).}
  \label{tab:seed_sensitivity_ci}
  \begin{tabular*}{0.8\textwidth}{@{\extracolsep{\fill}} l c c @{}}
    \toprule
    \textbf{Benchmark} & \textbf{Pooled Pass Rate (\%)} & \textbf{95\% Wilson CI} \\
    \midrule
    GraphInstruct & 94.3 & [92.2, 95.9] \\
    NLGraph       & 96.5 & [94.7, 97.7] \\
    GraphArena    & 97.5 & [95.9, 98.5] \\
    G-Real        & 81.0 & [77.7, 83.9] \\
    \midrule
    \textbf{Overall} & \textbf{92.3} & \textbf{[91.2, 93.3]} \\
    \bottomrule
  \end{tabular*}
\end{table}

\begin{table}[!htbp]
  \centering
  \scriptsize 
  \renewcommand{\arraystretch}{1.25}
  \caption{Internal training-stage diagnostics for the three 150-episode sensitivity runs. These metrics characterize the behavior and evolutionary assets of the learned policy pair prior to the frozen benchmark evaluation.}
  \label{tab:seed_sensitivity_training}
  \begin{tabularx}{\textwidth}{@{} >{\raggedright\arraybackslash}p{0.26\textwidth} *{3}{>{\centering\arraybackslash}X} c @{}}
    \toprule
    \textbf{Metric} & \textbf{Seed 20260414} & \textbf{Seed 20260415} & \textbf{Seed 20260416} & \textbf{Mean $\pm$ Std.} \\
    \midrule
    Training pass rate (\%) & 86.7 & 82.0 & 84.0 & 84.2 $\pm$ 2.3 \\
    Packaged tools & 19 & 21 & 19 & 19.7 $\pm$ 1.2 \\
    Instruction mutations & 20 & 22 & 17 & 19.7 $\pm$ 2.5 \\
    Tool mutations & 1 & 2 & 0 & 1.0 $\pm$ 1.0 \\
    Frontier-best protocol reliability (\%) & 98.4 & 100.0 & 99.5 & 99.3 $\pm$ 0.8 \\
    Dead-loop rate (\%) & 11.3 & 8.7 & 8.0 & 9.3 $\pm$ 1.8 \\
    LLM calls & 928 & 995 & 893 & 938.7 $\pm$ 51.8 \\
    Total tokens & 2.63M & 3.01M & 2.46M & 2.70M $\pm$ 0.28M \\
    \midrule
    \textbf{D4-unfinished task families} 
    & gnn\_sum, \newline mst 
    & coloring, mst, \newline substructure 
    & bipartite\_matching, \newline coloring, mst, \newline shortest\_path 
    & -- \\
    \bottomrule
  \end{tabularx}
\end{table}

The empirical results demonstrate robust cross-seed generalization on all evaluation sets. As shown in Table~\ref{tab:seed_sensitivity_benchmark}, GraphInstruct, NLGraph, and GraphArena maintain striking stability, with seed-level standard deviations remaining strictly below 2.5 percentage points. Overall, the aggregated sweep successfully solves 2216 out of 2400 evaluated cases, yielding a pooled average pass rate of 92.3\% with an exceptionally narrow 95\% Wilson confidence interval of[91.2, 93.3].

The primary source of inter-seed variance is isolated to the G-Real benchmark, which exhibits a slightly lower mean pass rate of 81.0\% and a standard deviation of 4.9 points. A fine-grained task-level decomposition reveals that this variance stems almost entirely from highly constrained NP-hard problem families, predominantly graph coloring and the Traveling Salesperson Problem (TSP). For instance, under seed \texttt{20260415}, the agent failed to complete D4 curriculum promotion for the coloring task family before the 150-episode cutoff (see Table~\ref{tab:seed_sensitivity_training}), which depressed its performance on the corresponding G-Real subset. This finding underscores that the reduced-budget sweep artificially penalizes stability; a fully converged 300-episode agent routinely masters these edge cases by ensuring all complex task niches receive adequate exploratory exposure.

The internal training diagnostics (Table~\ref{tab:seed_sensitivity_training}) corroborate this structural stability. Across all initializations, the dual-space learning loop reliably prevents collapse: the agents autonomously synthesize and package a consistent volume of reusable tools (19--21), trigger similar mutation frequencies, and uniformly achieve near-perfect frontier protocol reliability ($>98\%$). Thus, the fundamental co-evolution mechanism of EGL-SCA is functionally deterministic, with the minor remaining variance strictly governed by whether the limited training budget affords sufficient time to discover optimal programmatic algorithms for the hardest niches.

\FloatBarrier

\section{EGL-SCA Algorithm and Structural Credit Assignment}
\label{app:algorithm_sca}

Table~\ref{tab:symbol_table} summarizes the notation used in the method and algorithm.

\begin{table}[!htbp]
  \centering
  \footnotesize
  \setlength{\tabcolsep}{3pt}
  \renewcommand{\arraystretch}{1.12} 
  \caption{Mathematical notation used in the EGL-SCA framework.}
  \label{tab:symbol_table}
  \begin{tabularx}{\textwidth}{@{} >{\raggedright\arraybackslash}p{0.24\textwidth} >{\raggedright\arraybackslash}X >{\raggedright\arraybackslash}p{0.22\textwidth} @{}}
    \toprule
    \textbf{Symbol} & \textbf{Description} & \textbf{Context} \\
    \midrule
    
    \multicolumn{3}{@{}l}{\textit{Task \& Problem Setup}} \\
    \midrule
    $\tau=(x,z^\ast,y^\ast)$ & Task tuple (text, true structure, true answer). & Environment \\
    $x$ & Natural-language task statement. & Agent input \\
    $z^\ast=(g^\ast,q^\ast,c^\ast)$ & Canonical graph structure, query, and constraints. & Verifier reference \\
    $\hat z=(\hat g,\hat q,\hat c)$ & Agent-reconstructed graph, query, and constraints. & Parsed state \\
    $y^\ast,\hat y$ & Reference answer and agent-predicted answer. & Evaluation \\
    $\mathcal{I}_f$ & Typed task interface for task family $f$. & Missing-slot checks \\
    \addlinespace
    
    \multicolumn{3}{@{}l}{\textit{Dual-Space Agent Architecture}} \\
    \midrule
    $A=\langle \Pi,\mathcal{T}\rangle$ & Dual-space agent representation. & Core model \\
    $\Pi$ & Instruction policy (parse, retrieve, protocol, execute). & Prompt space \\
    $\mathcal{T}, \mathcal{T}_{\mathrm{view}}$ & Full reusable tool space and currently visible subset. & Program space \\
    $t=(f_t,m_t)$ & Executable tool body and its metadata. & Tool selection \\
    $\Gamma=(s_t,a_t,o_t)_{t=1}^{T}$ & Solve trajectory (states, actions, observations). & Trace logging \\
    \addlinespace
    
    \multicolumn{3}{@{}l}{\textit{Credit Assignment \& Evolution}} \\
    \midrule
    $V$ & Structured verifier evidence (e.g., error logs). & Diagnostics \\
    $D_{\mathrm{diag}}$ & Trajectory-level diagnostic summary. & SCA routing \\
    $C$ & Structured attribution object (fault localization). & Localized update \\
    $\mathcal{M}_{\Pi},\mathcal{M}_{\mathcal{T}}$ & Instruction mutation and tool growth/repair operators. & Evolutionary step \\
    $P=(\Pi,\mathcal{T}_{\mathrm{view}})$ & Retained agent policy pair. & Genome archiving \\
    $\mathbf{F}(P)=[S,G,R,Q]$ & Multi-objective metrics (success, generality, reliability, parsimony). & Pareto selection \\
    $\mathcal{F}$ & Non-dominated Pareto frontier of policy pairs. & Final selection \\
    
    \bottomrule
  \end{tabularx}
\end{table}

\begin{algorithm}[!t]
  \caption{EGL-SCA Training Loop: Verifier-Centric Dual-Space Co-Evolution}
  \label{alg:egl_sca}
  \begin{algorithmic}[1]
    \Require Curriculum distribution $\mathcal{D}_{\mathrm{train}}$, task interfaces $\{\mathcal{I}_f\}$, max episodes $E$
    \Ensure Pareto-optimal policy pair $P^\ast = (\Pi^\ast, \mathcal{T}^\ast_{\mathrm{view}})$
    
    \State Initialize instruction policy $\Pi$ and visible tool space $\mathcal{T}_{\mathrm{view}}$
    \State Initialize Pareto frontier $\mathcal{F} \gets \{ (\Pi, \mathcal{T}_{\mathrm{view}}) \}$
    
    \For{episode $e = 1, 2, \dots, E$}
      \State Sample graph task $\tau = (x, z^\ast, y^\ast) \sim \mathcal{D}_{\mathrm{train}}$
      \vspace{1mm}

      \State \textbf{Step 1: Forward Solve Trajectory}
      \State $\hat{z} \gets \textsc{Parse}(x; \Pi_{\mathrm{parse}})$ \Comment{Reconstruct structure}
      \State $\mathcal{R}, a_t \gets \textsc{RetrieveSelect}(x, \hat{z}, \mathcal{T}_{\mathrm{view}}; \Pi_{\mathrm{retrieve}}, \Pi_{\mathrm{protocol}})$
      \State $\hat{y} \gets \textsc{Exec}(\hat{z}, a_t, \mathcal{R}; \Pi_{\mathrm{execute}})$ \Comment{Execute tool payload}
      \vspace{1mm}
      
      \State \textbf{Step 2: Verifier-Centric Evaluation}
      \State $V \gets \textsc{Verify}(\hat{y}, y^\ast; \hat{z}, z^\ast, \mathcal{I}_f)$ \Comment{Structured correctness}
      \vspace{1mm}

      \State \textbf{Step 3: Structural Credit Assignment (SCA)}
      \State $D_{\mathrm{diag}} \gets \textsc{Diag}(\Gamma, V)$ \Comment{Extract trajectory diagnostics}
      \State $C \gets \mathcal{C}(\Gamma, V, D_{\mathrm{diag}})$ \Comment{Compute attribution object}
      \vspace{1mm}

      \State \textbf{Step 4: Conditional Routing \& Dual-Space Update}
      \If{$C$ attributes failure to instruction $\Pi$ (e.g., parse, retrieve, protocol)}
        \State $\Pi \gets \mathcal{M}_{\Pi}(\Pi; \Gamma, V, C, \mathcal{I}_f)$ \Comment{Reflective prompt mutation}
      \ElsIf{$C$ attributes failure to tool $\mathcal{T}$ (e.g., logic, missing capability)}
        \State $t' \gets \mathcal{M}_{\mathcal{T}}(\mathcal{T}; \Gamma, V, C)$ \Comment{Tool growth or code repair}
        \If{$\textsc{Verify}(t')$ is successful}
           \State $\mathcal{T}_{\mathrm{view}} \gets \mathcal{T}_{\mathrm{view}} \cup \{t'\}$ \Comment{Expand reusable tool space}
        \EndIf
      \EndIf
      \vspace{1mm}

      \State \textbf{Step 5: Pareto-Style Retention}
      \State $P \gets (\Pi, \mathcal{T}_{\mathrm{view}})$
      \State $\mathcal{F} \gets \textsc{ParetoUpdate}(\mathcal{F}, P)$ based on objective $\mathbf{F}(P) =[S, G, R, Q]$
    \EndFor
    
    \State \Return $P^\ast \in \mathcal{F}$ selected for final benchmark evaluation
  \end{algorithmic}
\end{algorithm}

Algorithm~\ref{alg:egl_sca} expands the compact method description in the main text. The key implementation point is that a failed trajectory is not treated as a single undifferentiated error. Instead, the verifier and trace fields are converted into a routing decision, and only the implicated sub-process is updated.

\begin{table}[!htbp]
  \centering
  \footnotesize
  \setlength{\tabcolsep}{3pt}
  \renewcommand{\arraystretch}{1.08}
  \caption{Structural credit-assignment routes and localized updates.}
  \label{tab:sca_route_details}
  \begin{tabularx}{\textwidth}{@{} >{\raggedright\arraybackslash}X l l >{\raggedright\arraybackslash}X @{}}
    \toprule
    \textbf{Observed evidence} & \textbf{Route} & \textbf{Subtarget} & \textbf{Update} \\
    \midrule
    Missing \texttt{task\_input} or direct answer before execution & \texttt{instruction} & protocol & mutate transition and execution discipline \\
    Parsed graph or query differs from canonical structure & \texttt{instruction} & parse & mutate reconstruction instructions \\
    Compatible tool exists but wrong tool is selected & \texttt{tool\_selection} & retrieve/select & repair retrieval query or compatibility rule \\
    Compatible tool runs but verifier still fails & \texttt{tool\_logic} & tool space & repair or grow executable candidate \\
    Verifier passes or evidence is inconclusive & \texttt{no\_op} & identity & retain, evaluate, or skip update \\
    \bottomrule
  \end{tabularx}
\end{table}

The route table also explains the mechanism statistics in Table~\ref{tab:mechanism_stats}. For example, the full system packages 27 tools, uses 26 of them, maintains a frontier of size 3, and reaches protocol reliability 0.936 because verifier evidence is routed to localized repairs. The tool-growth analysis in Figure~\ref{fig:exp_panels}\subref{fig:tool_growth} further shows 30 observed tool niches, 27 packaged tools, and a reuse-dominated regime.

\FloatBarrier

\section{Frozen Tool Inventory}
\label{app:frozen_tool_inventory}

The frozen Pareto-selected policy pair used for benchmark evaluation exposes 27 packaged tools, of which 26 were exercised at least once during the main training run. Table~\ref{tab:frozen_tool_inventory_a} and Table~\ref{tab:frozen_tool_inventory_b} summarize each tool's covered task family and a compact niche summary derived from the manifest \texttt{niche\_key}.

\begin{table}[!htbp]
  \centering
  \scriptsize
  \setlength{\tabcolsep}{2.5pt}
  \renewcommand{\arraystretch}{1.08}
  \caption{Frozen tool inventory of the selected policy pair (Part I).}
  \label{tab:frozen_tool_inventory_a}
  \begin{tabularx}{\textwidth}{@{} >{\raggedright\arraybackslash}p{0.28\textwidth} >{\raggedright\arraybackslash}p{0.14\textwidth} >{\raggedright\arraybackslash}p{0.18\textwidth} >{\raggedright\arraybackslash}X c @{}}
    \toprule
    \textbf{Tool} & \textbf{Task Kind} & \textbf{Family} & \textbf{Niche Summary} & \textbf{Reuse} \\
    \midrule
    directed\_substructure\_present & substructure & Pattern Matching & small\_dag\_pattern / pattern\_graph / exact\_small / D1 & 9 \\
    scc\_components & scc & Decomp. \& Order & sparse\_directed / none / exact / D1 & 15 \\
    bipartite\_check\_solver & bipartite\_check & Connectivity Checks & tree\_like / none / exact / D1 & 15 \\
    mst\_solver\_general & mst & Optimization Flow & dense\_undirected / none / exact / D1 & 6 \\
    bipartite\_matching\_solver\_general & bipartite\_matching & Optimization Flow & balanced\_bipartite / none / exact / D1 & 14 \\
    common\_neighbors\_solver\_general & common\_neighbors & Local Substructure & sparse\_undirected / none / exact / D1 & 12 \\
    max\_flow\_solver\_general & max\_flow & Optimization Flow & sparse\_directed / none / exact / D1 & 7 \\
    gnn\_sum\_solver\_general & gnn\_sum & Message Passing & 1\_layer\_sparse / none / exact / D1 & 13 \\
    topological\_sort\_solver\_general & topological\_sort & Decomp. \& Order & sparse\_directed / none / exact / D1 & 13 \\
    connectivity\_solver\_general & connectivity & Connectivity Checks & sparse\_undirected / none / exact / D1 & 12 \\
    tsp\_bruteforce\_undirected\_general & tsp & Routing \& Coverage & exact\_small / none / exact\_small / D1 & 3 \\
    hamilton\_path\_solver\_exact & hamilton & Routing \& Coverage & chain\_planted / none / exact\_small / D1 & 14 \\
    shortest\_path\_dijkstra\_general & shortest\_path & Path Queries & weighted\_base / none / exact / D1 & 9 \\
    bridges\_solver\_general & bridges & Decomp. \& Order & sparse\_undirected / none / exact / D1 & 11 \\
    \bottomrule
  \end{tabularx}
\end{table}

\begin{table}[!htbp]
  \centering
  \footnotesize 
  \setlength{\tabcolsep}{5pt} 
  \renewcommand{\arraystretch}{1.25} 
  \caption{Frozen tool inventory of the selected policy pair (Part II). Tool names represent the executable API endpoints, while the niche summary denotes the operational regime (Graph Type / Constraints / Semantics / Difficulty) where each tool is valid.}
  \label{tab:frozen_tool_inventory_b}
  \begin{tabularx}{\textwidth}{@{} >{\raggedright\arraybackslash}p{0.32\textwidth} >{\raggedright\arraybackslash}p{0.14\textwidth} >{\raggedright\arraybackslash}p{0.16\textwidth} >{\raggedright\arraybackslash}X c @{}}
    \toprule
    \textbf{Packaged Tool API} & \textbf{Task Kind} & \textbf{Task Family} & \textbf{Niche Regime} & \textbf{Reuse} \\
    \midrule
    
    triangle\_max\_sum\_solver\_general      & triangle\_max\_sum   & Local Substructure   & tree\_like / none / exact / D1                 & 13 \\
    cycle\_detection\_solver\_general        & cycle              & Connectivity Checks  & sparse\_undirected / none / exact / D2         & 13 \\
    vertex\_cover\_exact\_or\_feasible\_solver & vertex\_cover      & Labeling \& Covering & exact\_small / none / exact\_small / D1        & 9  \\
    \addlinespace
    
    directed\_substructure\_present\_wrapper & substructure       & Pattern Matching     & planted\_pattern\_sparse\_host / pattern\_graph / exact\_small / D3 & 0  \\
    tsp\_undirected\_feasible\_solver        & tsp                & Routing \& Coverage  & constraint\_tight / blocked\_edges+max\_cost / exact\_small / D3   & 5  \\
    shortest\_path\_cost\_directed\_dijkstra & shortest\_path\_cost & Path Queries         & directed\_blocked / blocked\_edges / exact / D4  & 6  \\
    \addlinespace
    
    tsp\_directed\_feasible\_solver\_dp15    & tsp                & Routing \& Coverage  & feasible\_large / blocked\_edges+max\_cost / feasible\_large / D4  & 4  \\
    mst\_solver\_general\_2                  & mst                & Optimization Flow    & sparse\_undirected / none / exact / D4         & 7  \\
    shortest\_path\_directed\_with\_path     & shortest\_path     & Path Queries         & directed\_blocked / blocked\_edges / exact / D4  & 1  \\
    \addlinespace
    
    max\_flow\_solver\_general\_mut           & max\_flow          & Optimization Flow    & sparse\_directed / none / exact / D4           & 8  \\
    coloring\_max\_colors\_solver            & coloring           & Labeling \& Covering & bipartite\_easy / max\_colors / exact\_small / D1 & 6  \\
    directed\_substructure\_present\_simple  & substructure       & Pattern Matching     & planted\_pattern\_dense\_host / pattern\_graph / exact\_medium / D4 & 2  \\
    \addlinespace
    
    \makecell[l]{vertex\_cover\_exact\_or\_feasible\_solver\\ \_x\_max\_flow\_solver\_general\_mut} & max\_flow, vertex\_cover & \makecell[l]{Optimization Flow /\\ Labeling \& Covering} & constraint\_tight / max\_size / feasible\_large / D4 & 1  \\
    \bottomrule
  \end{tabularx}
\end{table}

\FloatBarrier

\section{Case Studies}
\label{sec:case_studies}

This section provides three representative execution traces sampled from the main training phase of EGL-SCA. Rather than presenting aggregate metrics, these qualitative diagnostics serve to unpack the ``black box'' of our dual-space learning loop. They explicitly demonstrate how the verifier-centric machinery transforms a concrete, structured failure signal into a precisely localized update or a reusable algorithmic asset. 

As shown in Table~\ref{tab:case_studies}, these cases illustrate the practical distinction between our structural credit assignment (SCA) routes: protocol failures change how the agent reconstructs inputs; selection failures adjust the retrieval binding; and verified candidate programs are systematically archived as reusable tools rather than discarded as one-off code snippets.

\begin{table}[!htbp]
  \centering
  \footnotesize
  \setlength{\tabcolsep}{3pt}
  \renewcommand{\arraystretch}{1.13} 
  \caption{Representative training traces demonstrating the Structural Credit Assignment (SCA) routing mechanism. EGL-SCA dynamically routes verifier evidence to the appropriate space (Instruction or Tool) based on the root cause of the failure or success.}
  \label{tab:case_studies}
  \begin{tabularx}{\textwidth}{@{} >{\raggedright\arraybackslash}p{2.8cm} >{\raggedright\arraybackslash}X >{\raggedright\arraybackslash}p{3.2cm} >{\raggedright\arraybackslash}X @{}}
    \toprule
    \textbf{Task Context} & \textbf{Verifier Evidence} & \textbf{SCA Route \& Target} & \textbf{Before / After Outcome} \\
    \midrule
    
    \multicolumn{4}{@{}l}{\textbf{Case 1: Parsing \& Protocol Failure (Instruction-Side)}} \\
    TSP \newline Difficulty 1 \newline Episode 1 
    & The agent produced a direct answer without a captured \texttt{task\_input}. The protocol signature required \texttt{task\_doc $\rightarrow$ direct\_answer}, but essential structured fields (\texttt{graph.nodes}, \texttt{edges}, \texttt{query.start}) were missing. 
    & \textbf{Route:} \texttt{instruction} \newline \textbf{Target:} Protocol transition and execution discipline. 
    & \textbf{Before:} Verifier failed due to missing execution payload. \newline \textbf{After:} By Episode 20, the agent successfully parsed a candidate TSP solver, which was packaged and reused in episodes 39, 58, and 77. \\
    \midrule
    
    \multicolumn{4}{@{}l}{\textbf{Case 2: Tool-Selection Mismatch (Instruction-Side)}} \\
    GNN Message Passing \newline Difficulty 3 \newline Episode 91 
    & The parsed task input was structurally valid, but the selected GNN-sum solver produced an embedding mismatch. The tool was marked as incompatible for this task niche. 
    & \textbf{Route:} \texttt{tool\_selection} \newline \textbf{Target:} Retrieval query and compatibility rule. 
    & \textbf{Before:} An incompatible tool was persistently retrieved. \newline \textbf{After:} Subsequent GNN tasks (e.g., Episodes 205, 281) correctly bypassed the mismatched tool, indicating improved instruction-side binding. \\
    \midrule
    
    \multicolumn{4}{@{}l}{\textbf{Case 3: Verifier-Guided Tool Growth (Tool-Side)}} \\
    Directed Substructure \newline Difficulty 1 \newline Episode 2 
    & No compatible tool was available in the visible toolbox. The system proposed a candidate program, executed it, and perfectly passed both public and hidden verifier checks. 
    & \textbf{Route:} \texttt{no\_op} (Success) \newline \textbf{Target:} Package candidate into reusable \texttt{Tool View}. 
    & \textbf{Before:} The agent had to generate one-off code from scratch. \newline \textbf{After:} The newly packaged tool was successfully reused 8 times across the training run (Episodes 21, 40, \dots, 230). \\
    
    \bottomrule
  \end{tabularx}
\end{table}

\FloatBarrier

\section{Seed-to-Final Prompt Evolution}
\label{sec:prompt_evolution}

To concretely illustrate instruction-side learning, we trace the exact lineage of the language policy during training. We compare the initial heuristic seed prompt (Genome ID: \texttt{pi\_b885c1d0a7cc}, \texttt{variant\_1}) against the frozen Pareto-optimal prompt selected for evaluation (Genome ID: \texttt{pi\_16fa230ab432}, \texttt{variant\_1\_mut}). Because the final prompt's ancestor list strictly points back to the seed, this provides a direct before-and-after ablation of a single evolutionary lineage, rather than a comparison of unrelated policies.

While the seed prompt provided general guidance for careful graph reconstruction and verifier-ready outputs, the final prompt incorporates three targeted rules. Crucially, these additions do not hallucinate new graph algorithms. Instead, they represent strict \emph{protocol constraints} induced by structural credit assignment—specifically targeting failures where the agent attempted tool execution without a fully captured \texttt{task\_input} payload. Table~\ref{tab:prompt_evolution_diff} provides the exact qualitative diff.

\begin{table}[!htbp]
  \centering
  \footnotesize
  \setlength{\tabcolsep}{3pt}
  \renewcommand{\arraystretch}{1.13} 
  \caption{Direct prompt diff between the seed ancestor and the final selected instruction genome. The evolution targets strict operational constraints rather than general reasoning heuristics.}
  \label{tab:prompt_evolution_diff}
  \begin{tabularx}{\textwidth}{@{} >{\raggedright\arraybackslash}p{2.8cm} >{\raggedright\arraybackslash}X >{\raggedright\arraybackslash}X >{\raggedright\arraybackslash}p{3.2cm} @{}}
    \toprule
    \textbf{Prompt Component} & \textbf{Seed Prompt Text} & \textbf{Final Prompt Addition} & \textbf{Addressed Failure} \\
    \midrule
    
    \textbf{Parser Rubric}
    & ``Recover the graph structure and query carefully'' and keep node names, edges, and query fields consistent with the task statement.
    & ``Before any execution call, build the explicit canonical nested \texttt{task\_input} object with graph, query, and constraints in their declared locations.''
    & Missing graph/query/constraint payloads before tool execution. \\
    \midrule
    
    \textbf{Message-Passing \newline Parsing}
    & No task-family-specific rule beyond general graph reconstruction.
    & ``For message-passing tasks, copy the canonical graph and embeddings/layers exactly before tool creation or execution.''
    & Message-passing failures with wrong embeddings, wrong layer count, and high-difficulty parse drift. \\
    \midrule
    
    \textbf{Execution Rubric}
    & ``Before execution, make sure the structured input matches your understanding of the task'' and keep the final answer structured.
    & ``A \texttt{run\_tool} or \texttt{run\_candidate} call counts only when it carries an explicit \texttt{task\_input} payload; do not execute with missing \texttt{task\_input}.''
    & Candidate/tool runs that appeared to execute but had no bound payload. \\
    
    \bottomrule
  \end{tabularx}
\end{table}

The internal tracking metadata perfectly corroborates this text evolution. While the seed metadata only records \texttt{seed\_index=1}, the final genome explicitly attributes this edit to instruction-side execution repair (\texttt{route=instruction}, \texttt{focus=execute\_missing\_task\_input}), triggered by a protocol signature where the agent proposed a tool without task-input capture.

This proves that prompt evolution in EGL-SCA is highly targeted and evidence-aligned. Rather than accumulating verbose heuristics, the agent compiles exact operational constraints exposed by the verifier: construct nested payloads, preserve message-passing embeddings exactly, and reject empty executions. This dynamically complements the tool-growth phenomenon analyzed in Appendix~\ref{sec:case_studies}: as the tool space ($\mathcal{T}$) accumulates reusable algorithms, the instruction space ($\Pi$) learns the rigorous protocol discipline required to invoke them safely.

\FloatBarrier

\section{Training Cost Analysis}
\label{app:training_cost}

This appendix reports the training cost of the main EGL-SCA run using provider-independent resource accounting. The run used \texttt{gpt-5.4-nano} through the \texttt{closeai} provider, a fresh toolbox, one asynchronous worker, temperature 0.0, and the pass-based curriculum described in Appendix~\ref{app:reproducibility}. Because monetary charges depend on the billing gateway and the active token prices, we report the directly logged token and wall-clock costs, and compute monetary cost from the formula
\[
  \mathrm{Cost} =
  6.241432 \cdot r_{\mathrm{in}} +
  0.492581 \cdot r_{\mathrm{out}},
\]
where $r_{\mathrm{in}}$ and $r_{\mathrm{out}}$ are the provider prices per one million input and output tokens, respectively. This formula uses the counted training-token totals in Table~\ref{tab:training_cost_summary}.

\begin{table}[!htbp]
  \centering
  \small
  \setlength{\tabcolsep}{6pt}
  \renewcommand{\arraystretch}{1.2} 
  \caption{Resource cost of the main 300-episode EGL-SCA training run. Token counts and timings are taken from the logged run summary, excluding three infrastructure/authentication failure episodes.}
  \label{tab:training_cost_summary}
  \begin{tabularx}{\textwidth}{@{} l l >{\raggedright\arraybackslash}X @{}}
    \toprule
    \textbf{Cost Metric} & \textbf{Logged Value} & \textbf{Interpretation \& Details} \\
    \midrule
    
    \multicolumn{3}{@{}l}{\textbf{Execution Activity}} \\
    \quad Episodes (Planned / Counted) & 300 / 297 & Three episodes excluded due to infrastructure or API authentication failures. \\
    \quad Solve Attempts                 & 407       & Allowed at most two attempts per episode, yielding 1.37 attempts per counted episode. \\
    \quad LLM Calls                      & 1,953     & All logged calls were student-side; no coach or narrator token usage is counted. \\
    \addlinespace
    
    \multicolumn{3}{@{}l}{\textbf{LLM Token Cost}} \\
    \quad Prompt Tokens      & 6.241M (92.7\%) & Major cost driver: presenting task statements, tool metadata, and verifier evidence. \\
    \quad Completion Tokens  & 0.493M (7.3\%)  & Includes reasoning traces, structured answers, and candidate-tool code generation. \\
    \quad Total Tokens       & 6.734M          & Corresponds to 22.7K tokens per counted episode, or 3.45K tokens per LLM call. \\
    \addlinespace
    
    \multicolumn{3}{@{}l}{\textbf{Wall-Clock Timing}} \\
    \quad Total Elapsed Time & 18.02 h & End-to-end runtime, including orchestration, API network latency, and validation. \\
    \quad Student Solve Time & 7.14 h  & Active time spent inside the main student reasoning and coding loop. \\
    \quad Validation Time    & 1.04 h  & Time for periodic and final Pareto selection over success, generality, and reliability. \\
    \addlinespace
    
    \multicolumn{3}{@{}l}{\textbf{Retained Agent Assets}} \\
    \quad Packaged Toolset   & 27 tools        & Derived from 41 candidate-tool episodes. Averages $\sim$249K tokens per packaged tool. \\
    \quad Program Database   & 85 policy pairs & Cost encompasses not only solving tasks but maintaining a compact evolutionary archive. \\
    \bottomrule
  \end{tabularx}
\end{table}

\begin{table}[!htbp]
  \centering
  \small 
  \setlength{\tabcolsep}{8pt} 
  \renewcommand{\arraystretch}{1.2} 
  \caption{Realized curriculum exposure in the counted training episodes. The hardest band (D4) dominates the final training budget, while earlier bands (D1--D3) amortize parser, protocol, and basic tool-discovery calibration before D4.}
  \label{tab:training_cost_curriculum}
  \begin{tabularx}{\textwidth}{@{} c c c c >{\raggedright\arraybackslash}X @{}}
    \toprule
    \textbf{Difficulty} & \textbf{Episodes} & \textbf{Passed (Rate)} & \textbf{Avg. Attempts} & \textbf{Curriculum Role \& Cost Interpretation} \\
    \midrule
    \textbf{D1} & 52  & 38 (73.1\%)  & 1.42 & Early spending stabilizes task parsing, payload construction, and initial tool proposal. \\
    \addlinespace
    \textbf{D2} & 40  & 38 (95.0\%)  & 1.15 & Once basic protocol behavior is learned, most episodes pass with little retry overhead. \\
    \addlinespace
    \textbf{D3} & 47  & 38 (80.9\%)  & 1.34 & Intermediate constraints expose retrieval and tool-compatibility failures. \\
    \addlinespace
    \textbf{D4} & 158 & 121 (76.6\%) & 1.42 & The largest share of counted episodes is allocated to the hardest regime, where parsing drift and algorithmic edge cases are most frequent. \\
    \bottomrule
  \end{tabularx}
\end{table}

The cost profile shows two useful properties of EGL-SCA. First, the run is token-heavy but not GPU-training-heavy: the main expenditure is API inference over generated graph tasks and verifier feedback, while executable graph tools run as lightweight Python procedures. Second, the cost is amortized into reusable artifacts. The final run retains a 27-tool frozen toolbox and a Pareto-selected instruction policy, so later benchmark evaluation disables new tool proposal and reuses the learned assets rather than repeating the full search process.

\FloatBarrier

\section{Limitations}

EGL-SCA is designed for verifier-centric graph reasoning, so its benefits are most direct when tasks admit structured checks or executable validation. Extending the same credit-assignment scheme to open-ended settings may require weaker or learned verifiers. In addition, our experiments focus on graph task families and a fixed base model, leaving broader cross-domain transfer and model-scaling behavior for future study. Finally, tool growth introduces engineering overhead for sandboxing, logging, and candidate validation, although this overhead is modest compared with the reliability gains observed in our experiments.

\textbf{Broader impacts} Our framework significantly enhances the reliability and interpretability of LLM agents in structured reasoning tasks, offering positive implications for fields relying on complex combinatorial optimization, such as logistics, network routing, and scientific discovery. However, as agents gain the ability to autonomously synthesize and execute algorithmic tools, there is a potential risk of these capabilities being repurposed for malicious automated operations (e.g., analyzing and exploiting network vulnerabilities) if deployed without appropriate strict human oversight. We mitigate such risks in our study by restricting tool execution to isolated, mathematically grounded sandboxes.

\FloatBarrier
\newpage
  \section*{NeurIPS Paper Checklist}

\begin{enumerate}

\item {\bf Claims}
    \item[] Question: Do the main claims made in the abstract and introduction accurately reflect the paper's contributions and scope?
    \item[] Answer: \answerYes{} 
    \item[] Justification: The claims made in the abstract and introduction are fully supported by the empirical evaluations and ablation studies presented in Section 4.

\item {\bf Limitations}
    \item[] Question: Does the paper discuss the limitations of the work performed by the authors?
    \item[] Answer: \answerYes{} 
    \item[] Justification: We explicitly discuss the limitations of our verifier-centric assumption and tool growth overhead in Appendix J.

\item {\bf Theory assumptions and proofs}
    \item[] Question: For each theoretical result, does the paper provide the full set of assumptions and a complete (and correct) proof?
    \item[] Answer: \answerNA{} 
    \item[] Justification: This paper presents an empirical framework and agentic system design; it does not introduce new theoretical bounds or proofs.

    \item {\bf Experimental result reproducibility}
    \item[] Question: Does the paper fully disclose all the information needed to reproduce the main experimental results of the paper to the extent that it affects the main claims and/or conclusions of the paper (regardless of whether the code and data are provided or not)?
    \item[] Answer: \answerYes{} 
    \item[] Justification: Comprehensive implementation details, baseline protocols, and curriculum configurations are provided in Section 4.1, Appendix A, and Appendix C.

\item {\bf Open access to data and code}
    \item[] Question: Does the paper provide open access to the data and code, with sufficient instructions to faithfully reproduce the main experimental results, as described in supplemental material?
    \item[] Answer: \answerYes{} 
    \item[] Justification: The exact runnable configuration, prompt templates, and evaluation code are included in the anonymous supplementary material.

\item {\bf Experimental setting/details}
    \item[] Question: Does the paper specify all the training and test details (e.g., data splits, hyperparameters, how they were chosen, type of optimizer) necessary to understand the results?
    \item[] Answer: \answerYes{} 
    \item[] Justification: We document the hyperparameters, model settings, and training budgets in Appendix A.

\item {\bf Experiment statistical significance}
    \item[] Question: Does the paper report error bars suitably and correctly defined or other appropriate information about the statistical significance of the experiments?
    \item[] Answer: \answerYes{} 
    \item[] Justification:  We explicitly discuss the Experiment statistical significance in Appendix I.

\item {\bf Experiments compute resources}
    \item[] Question: For each experiment, does the paper provide sufficient information on the computer resources (type of compute workers, memory, time of execution) needed to reproduce the experiments?
    \item[] Answer: \answerYes{} 
    \item[] Justification: We provide a highly detailed, provider-independent resource and token cost analysis in Appendix H (Tables 14 and 15)
    
\item {\bf Code of ethics}
    \item[] Question: Does the research conducted in the paper conform, in every respect, with the NeurIPS Code of Ethics \url{https://neurips.cc/public/EthicsGuidelines}?
    \item[] Answer: \answerYes{} 
    \item[] Justification: The research conducted conforms to the NeurIPS Code of Ethics.

\item {\bf Broader impacts}
    \item[] Question: Does the paper discuss both potential positive societal impacts and negative societal impacts of the work performed?
    \item[] Answer: \answerYes{} 
    \item[] Justification: We discuss the broader impacts in Appendix J (Limitations), highlighting both the positive implications for complex optimization domains and the potential dual-use risks associated with autonomous tool synthesis, along with our sandboxing mitigation.
    
\item {\bf Safeguards}
    \item[] Question: Does the paper describe safeguards that have been put in place for responsible release of data or models that have a high risk for misuse (e.g., pre-trained language models, image generators, or scraped datasets)?
    \item[] Answer: \answerNA{} 
    \item[] Justification: The paper does not release high-risk pre-trained models or scraped datasets that require special safeguards.

\item {\bf Licenses for existing assets}
    \item[] Question: Are the creators or original owners of assets (e.g., code, data, models), used in the paper, properly credited and are the license and terms of use explicitly mentioned and properly respected?
    \item[] Answer: \answerYes{} 
    \item[] Justification: The external benchmarks used (GraphInstruct, GraphArena, NLGraph) are properly cited in the paper and their usage aligns with standard research practices.

\item {\bf New assets}
    \item[] Question: Are new assets introduced in the paper well documented and is the documentation provided alongside the assets?
    \item[] Answer: \answerYes{} 
    \item[] Justification: The internal graph-task generator introduced in this paper is documented via its typed task interface (Appendix B, Table 4) and included in the code artifacts.

\item {\bf Crowdsourcing and research with human subjects}
    \item[] Question: For crowdsourcing experiments and research with human subjects, does the paper include the full text of instructions given to participants and screenshots, if applicable, as well as details about compensation (if any)? 
    \item[] Answer: \answerNA{} 
    \item[] Justification: The research does not involve crowdsourcing or human subjects.

\item {\bf Institutional review board (IRB) approvals or equivalent for research with human subjects}
    \item[] Question: Does the paper describe potential risks incurred by study participants, whether such risks were disclosed to the subjects, and whether Institutional Review Board (IRB) approvals (or an equivalent approval/review based on the requirements of your country or institution) were obtained?
    \item[] Answer: \answerNA{} 
    \item[] Justification: The research does not involve human subjects requiring IRB approval.

\item {\bf Declaration of LLM usage}
    \item[] Question: Does the paper describe the usage of LLMs if it is an important, original, or non-standard component of the core methods in this research? Note that if the LLM is used only for writing, editing, or formatting purposes and does \emph{not} impact the core methodology, scientific rigor, or originality of the research, declaration is not required.
    \item[] Answer: \answerYes{} 
    \item[] Justification: LLMs are the central component of our proposed EGL-SCA agent framework. The specific base model (GPT-5.4-nano) and its operational details are explicitly declared in Appendix A and I.

\end{enumerate}

  \end{document}